%% file: bare_jrnl_new_sample4.tex
\documentclass[lettersize,journal]{IEEEtran}

\usepackage{graphicx}
\usepackage{multirow}
\usepackage{hyperref}       
\usepackage{url}            
\usepackage{booktabs}       
\usepackage{amsfonts}       
\usepackage{nicefrac}       
\usepackage{microtype}      
\usepackage{xcolor}         

\usepackage{mathrsfs}
\usepackage{amsmath}

\usepackage{bm}
\usepackage{algorithm}
\usepackage{algorithmic}
\usepackage{float}  
\usepackage{lipsum}

\usepackage{xcolor}    
\usepackage{varwidth}  

\usepackage{amssymb}
\usepackage{pifont}
 \usepackage{enumitem}
\usepackage{multirow}
\usepackage{amsthm}
\theoremstyle{definition}

\usepackage{wrapfig}

\usepackage{booktabs}  
\usepackage{array}     
\usepackage{graphicx}  
\usepackage{subfigure}
\usepackage{multirow}  
\usepackage{multicol}  

\usepackage{colortbl}
\usepackage{xcolor}
\usepackage{array}
\usepackage{xspace}
\newcommand{\ourmethod}{\textsc{{MPhil}}\xspace}

\begin{document}

\title{Raising the Bar in Graph OOD Generalization: Invariant Learning Beyond Explicit Environment Modeling}

\author{Xu Shen, Yixin Liu, Yili Wang, Rui Miao, Yiwei Dai, Shirui Pan, Yi Chang and Xin Wang
 \thanks{Xu Shen (shenxu23@mails.jlu.edu.cn), Yili Wang (wangyl21@mails.jlu.edu.cn), Rui Miao (ruimiao20@mails.jlu.edu.cn), Yiwei Dai (daiyw23@mails.jlu.edu.cn), Yi chang and Xin Wang (yichang@jlu.edu.cn, xinwang@jlu.edu.cn) are with The School of Artificial Intelligence, Jilin University, Changchun, China.}
 \thanks{Yixin Liu (yixin.liu@griffith.edu.au) and Shirui Pan (s.pan@griffith.edu.au) with The School of Information and Communication Technology, Griffith University, Australia.}
 \thanks{Xu Shen and Yixin Liu are equal contribution.}
 \thanks{Xin Wang is the corresponding author.}
}
\markboth{Journal of \LaTeX\ Class Files,~Vol.~14, No.~8, August~2021}%
{Shell \MakeLowercase{\textit{et al.}}: A Sample Article Using IEEEtran.cls for IEEE Journals}


\maketitle

\begin{abstract}
Out-of-distribution (OOD) generalization has emerged as a critical challenge in graph learning, as real-world graph data often exhibit diverse and shifting environments that traditional models fail to generalize across. A promising solution to address this issue is graph invariant learning (GIL), which aims to learn invariant representations by disentangling label-correlated invariant subgraphs from environment-specific subgraphs. However, existing GIL methods face two major challenges: (1) the difficulty of \textbf{capturing and modeling diverse environments} in graph data, and (2) the \textbf{semantic cliff}, where invariant subgraphs from different classes are difficult to distinguish, leading to poor class separability and increased misclassifications. 
To tackle these challenges, we propose a novel method termed \textbf{M}ulti-\textbf{P}rototype \textbf{H}yperspherical \textbf{I}nvariant \textbf{L}earning (\ourmethod), which introduces two key innovations: (1) \textit{hyperspherical invariant representation extraction}, enabling robust and highly discriminative hyperspherical invariant feature extraction, and (2) \textit{multi-prototype hyperspherical classification}, which employs class prototypes as intermediate variables to eliminate the need for explicit environment modeling in GIL and mitigate the semantic cliff issue. Derived from the theoretical framework of GIL, we introduce two novel objective functions: the \textit{invariant prototype matching loss} to ensure samples are matched to the correct class prototypes, and the \textit{prototype separation loss} to increase the distinction between prototypes of different classes in the hyperspherical space.
Extensive experiments on 13 OOD generalization benchmark datasets demonstrate that \ourmethod achieves state-of-the-art performance, significantly outperforming existing methods across graph data from various domains and with different distribution shifts. The source code of \ourmethod is available at \href{https://anonymous.4open.science/r/MPHIL-7D9E/}{https://anonymous.4open.science/r/MPHIL-7D9E/}.
\end{abstract}


\section{Introduction} \label{sec:intro}
\input{1_intro/intro_v3}
\section{Related Works} \label{appe:rw}
\input{6_appendix/rw}
\section{Preliminaries and Background}
\input{2_preliminary/pre_v2}

\section{Methodology}
\input{3_method/method_v2}

\section{Experiments}
\input{4_exp/exp_v2}

\section{Conclusion}
\input{5_conclusion/conclusion}
\bibliographystyle{IEEEtran}
\bibliography{IEEEabrv,bare_jrnl_new_sample4}

\newpage

\begin{IEEEbiography}[{\includegraphics[width=1in,height=1.25in, clip,keepaspectratio]{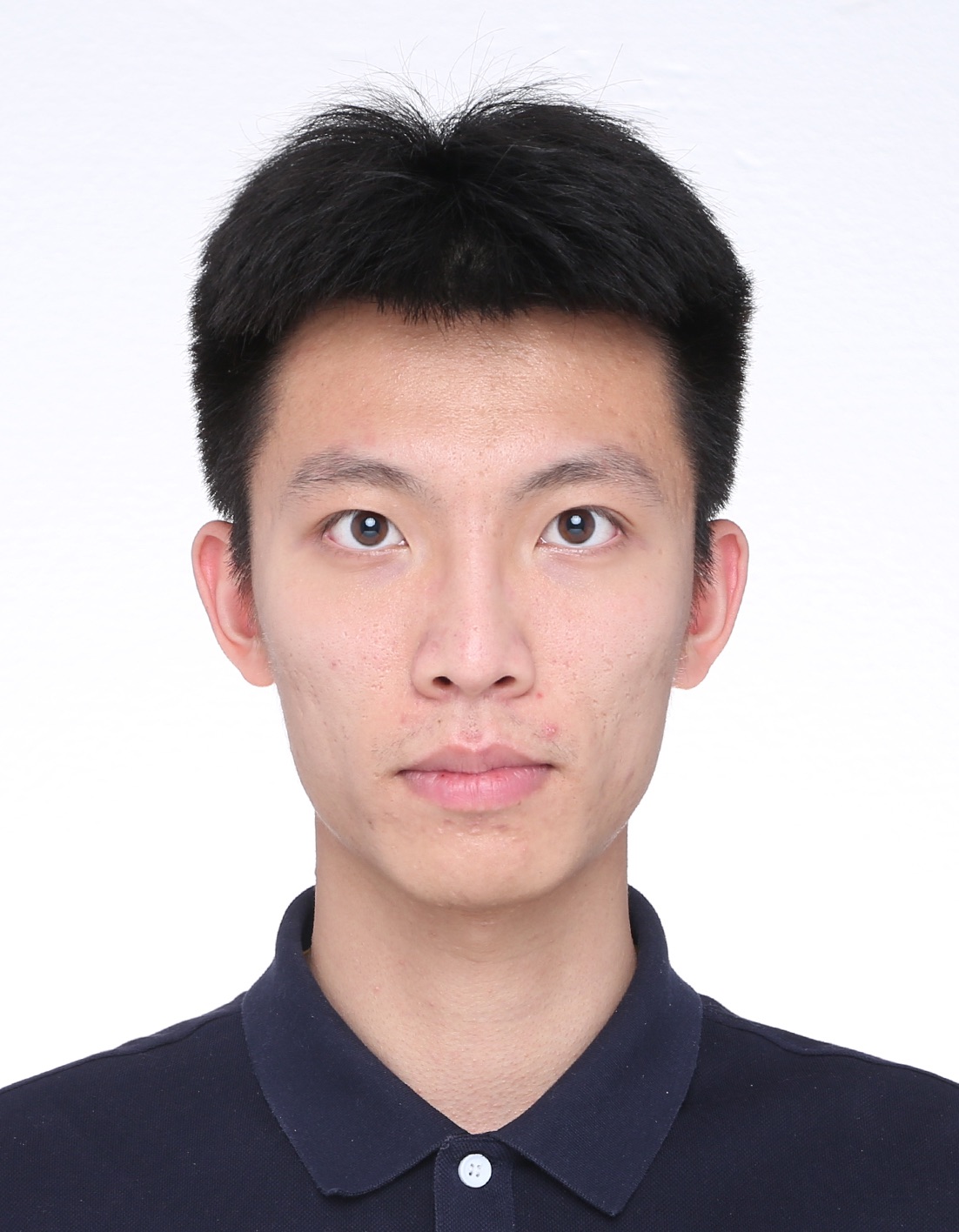}}]{Xu Shen} is currently a  master degree student in the School of Artificial Intelligence at Jilin University, supervised by Associate Professor Xin Wang. His research interests include graph representation learning, out-of-distribution generalization and large language models on graphs.
\end{IEEEbiography}
\vspace{-10pt}

\begin{IEEEbiography}[{\includegraphics[width=1in,height=1.25in,clip,keepaspectratio]{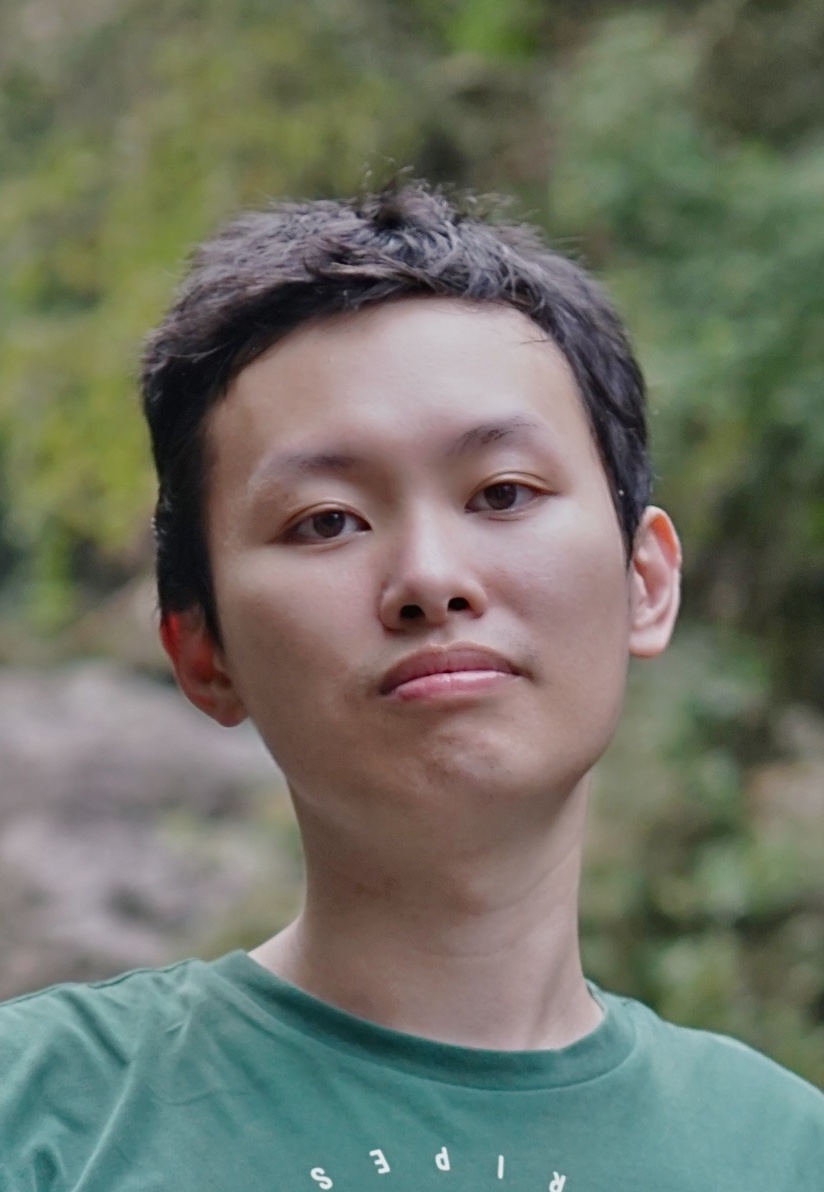}}]{Yixin Liu}
is a postdoctoral research fellow at Griffith University. He received his Ph.D. in Artificial Intelligence (AI) from Monash University, Australia, and his Bachelor’s and Master's degrees from Beihang University, China. His research concentrates on data mining, machine learning, graph analytics, and anomaly detection. To date, he has published more than 20 research papers in top-tier journals and conferences, including IEEE TKDE, IEEE TNNLS, NeurIPS, KDD, AAAI, and Web Conference. He is a recipient of Google Ph.D. Fellowship in 2022.
\end{IEEEbiography}
\vspace{-10pt}

\begin{IEEEbiography}[{\includegraphics[width=1in,height=1.25in, clip,keepaspectratio]{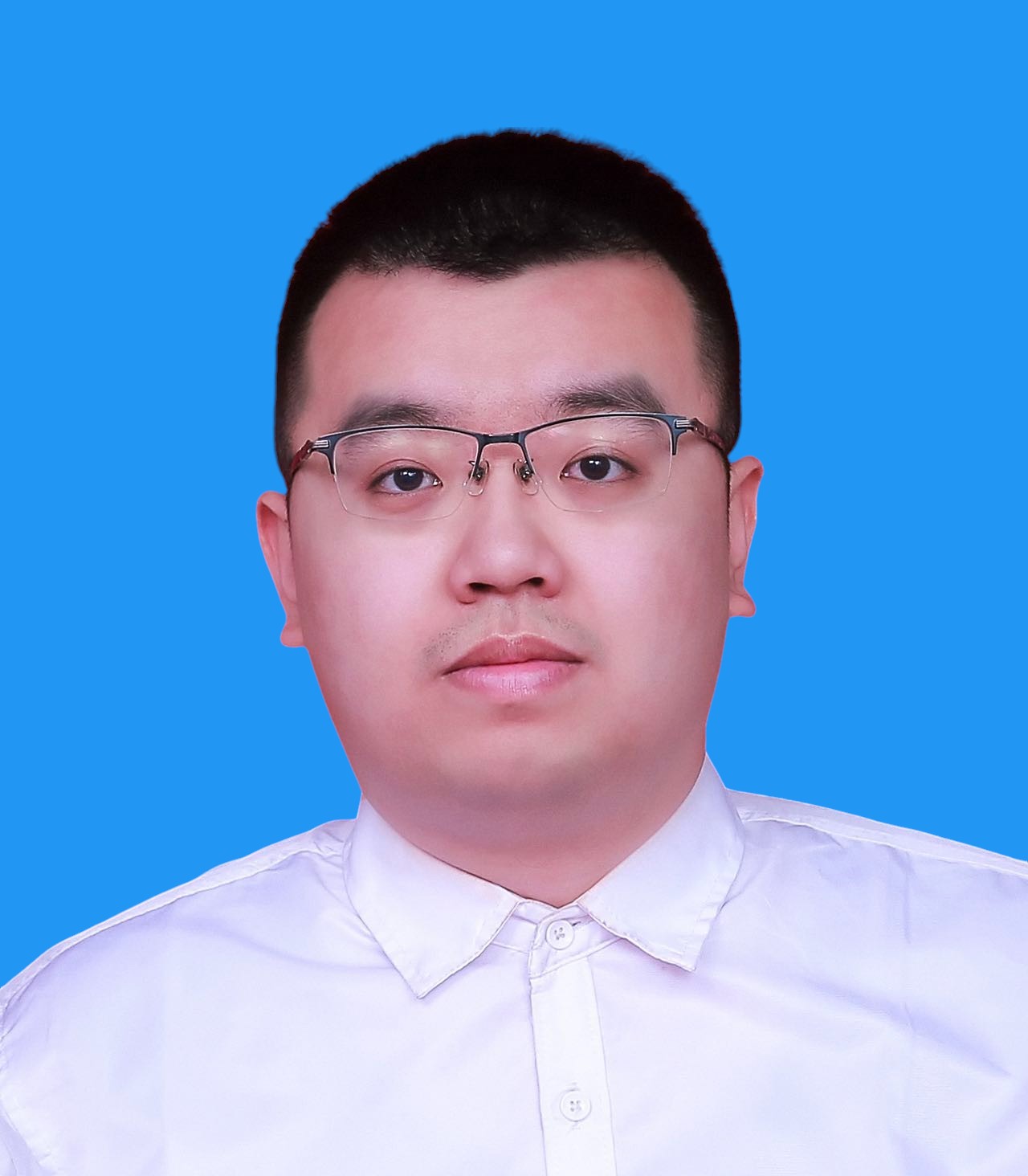}}]{Yili Wang} is a postdoctoral researcher at the School of Artificial Intelligence, Jilin University, supervised by Professor Yi Chang.  He received his doctorate from the School of Artificial Intelligence, Jilin University. He received his Master's degree from the School of Software, Jilin University. His research focuses on large-scale graph representation learning, including node classification, molecular graph classification, graph OOD detection, and generalization. He specializes in techniques such as contrastive learning, sharpness-aware minimization, adversarial learning, and diffusion models.
\end{IEEEbiography}
\vspace{-10pt}

\begin{IEEEbiography}[{\includegraphics[width=1in,height=1.25in,clip,keepaspectratio]{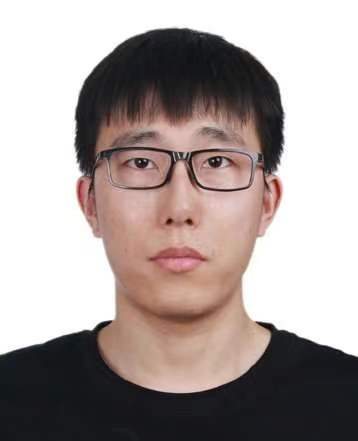}}]{Rui Miao} is a Ph.D. candidate at the School of Artificial Intelligence, Jilin University, supervised by Associate Professor Xin Wang. He received his Master's degree from the School of Artificial Intelligence, Jilin University. His research focuses on large-scale graph representation learning, including node classification, link prediction, and robustness of graphs. He specializes in techniques such as graph representation learning, self-supervised learning, and trustworthy machine learning.
\end{IEEEbiography}
\vspace{-10pt}

\begin{IEEEbiography}[{\includegraphics[width=1in,height=1.25in, clip,keepaspectratio]{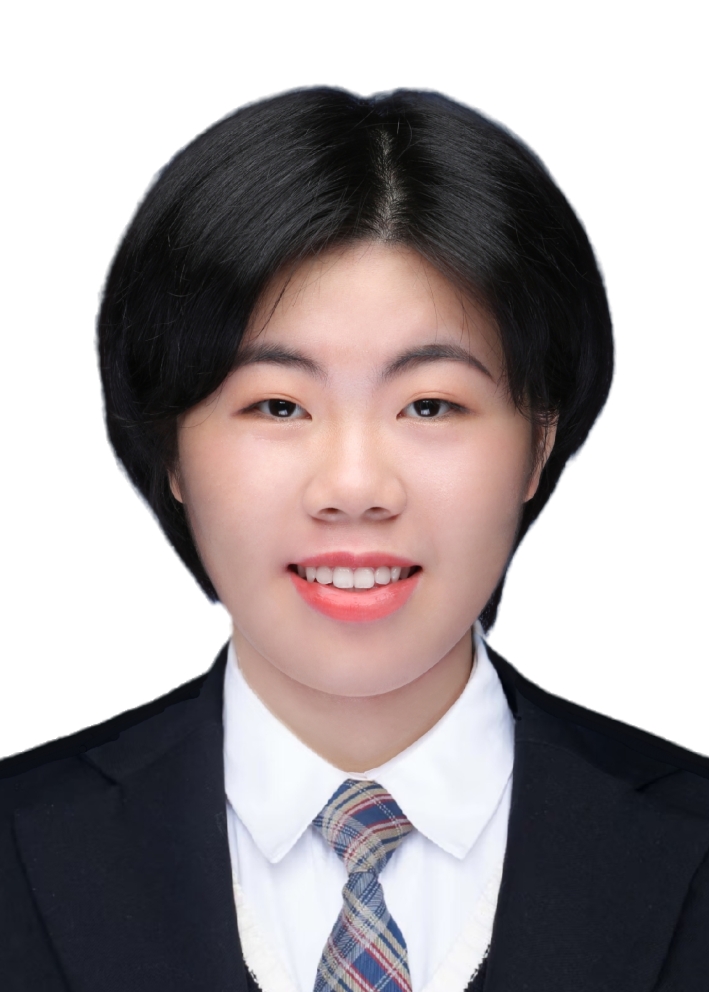}}]{Yiwei Dai} is a Master's student in the School of Artificial Intelligence at Jilin University, China. She received the B.E. degree in Computer Science and Technology from Jilin University in 2023. Her research interests include fairness in large language models, the mitigation of social biases, and post-training techniques for foundation models.
\end{IEEEbiography}
\vspace{-10pt}

\begin{IEEEbiography}[{\includegraphics[width=1in,height=1.25in,clip,keepaspectratio]{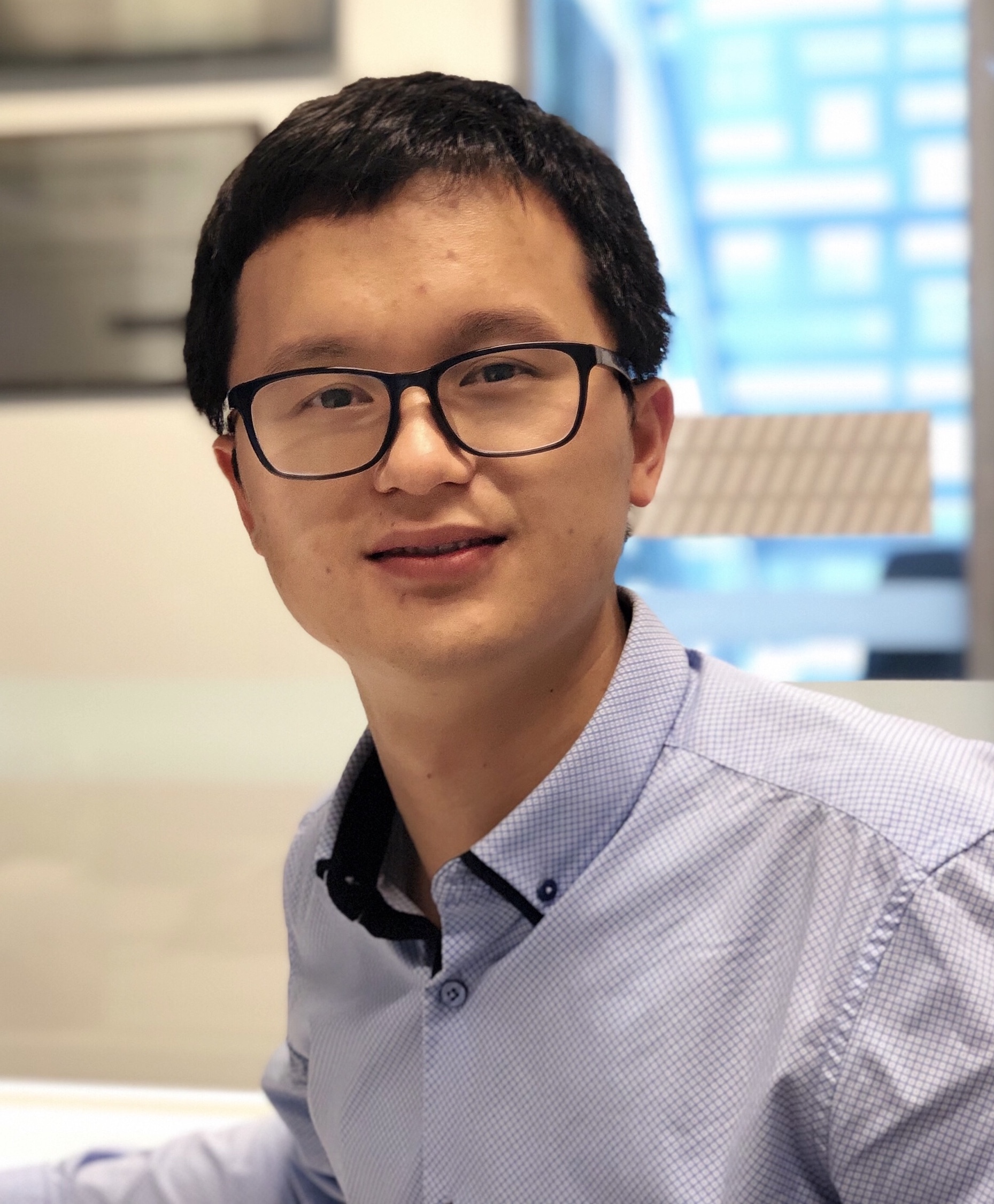}}]{Shirui Pan} received his Ph.D. in Computer Science from the University of Technology Sydney (UTS) and is a Professor in the School of Information and Communication Technology at Griffith University, Australia. 
His research focus on data mining and machine learning and has been published in top venues, including Nature Machine Intelligence, KDD, and ICLR, among others. He has received several prestigious awards, including the 2024 IEEE CIS TNNLS Oustanding Paper Award, the 2020 IEEE ICDM Best Student Paper Award, the 2024 AI's 10 to Watch recognition, and the 2024 IEEE ICDM Tao Li Award. He is also an ARC Future Fellow.
\end{IEEEbiography}
\vspace{-10pt}

\begin{IEEEbiography}[{\includegraphics[width=1in,height=1.25in,clip,keepaspectratio]{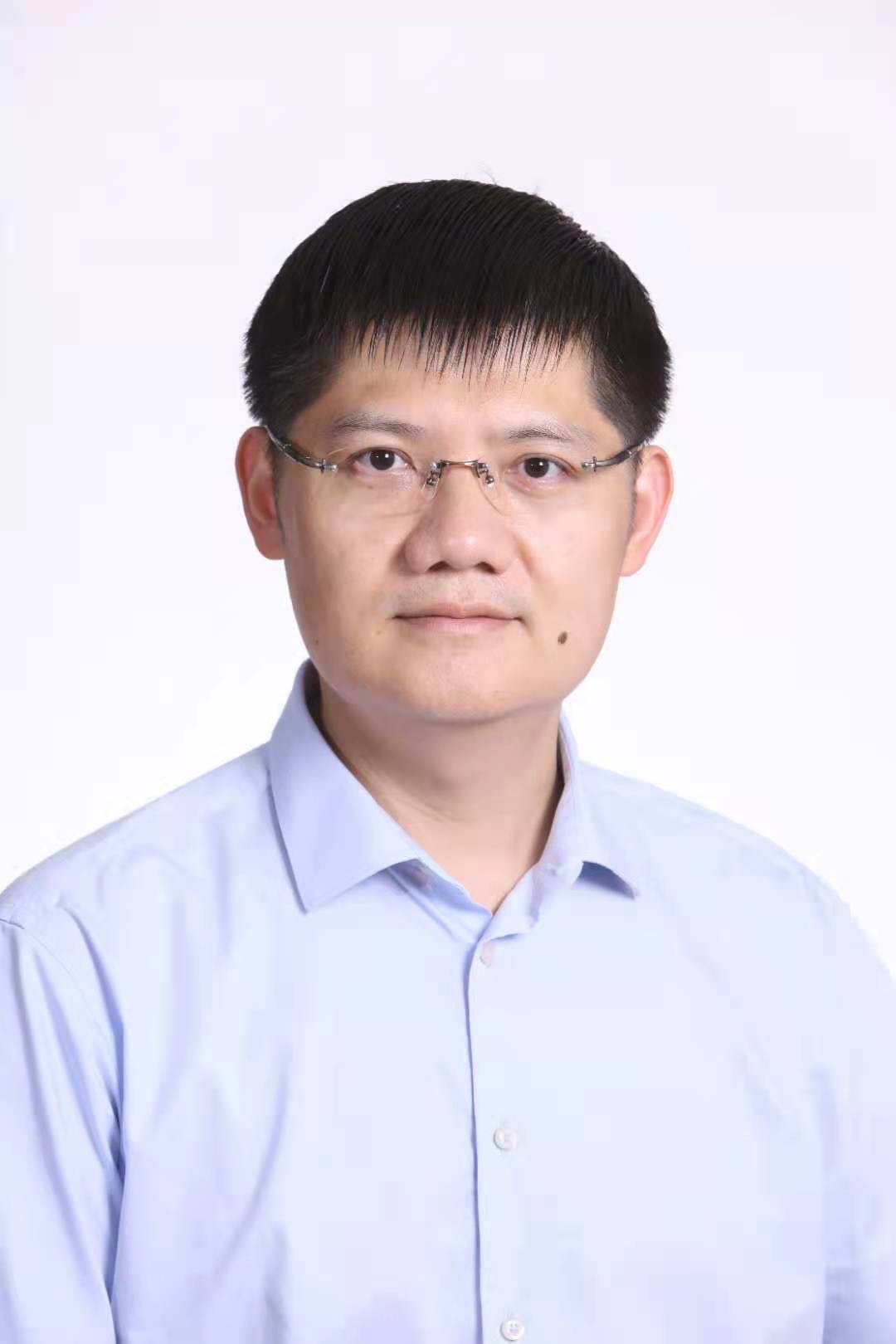}}]{Yi Chang} is the dean with the School of Artificial Intelligence, Jilin University, Changchun, China. He was elected as a Chinese National distinguished professor, in 2017 and an ACM distinguished scientist, in 2018. Before joining academia, he was the technical vice president with Huawei Research America, and the research director with Yahoo Labs. He is the author of two books and more than 100 papers in top conferences or journals. His research interests include information retrieval, data mining, machine learning, natural language processing, and artificial intelligence. He won the Best Paper Award on KDD’2016
and WSDM’2016.He has served as a Conference general chair for WSDM’2018, SIGIR’2020 and WWW’2025. 
\end{IEEEbiography}
\vspace{-10pt}

\begin{IEEEbiography}[{\includegraphics[width=1in,height=1.25in,clip,keepaspectratio]{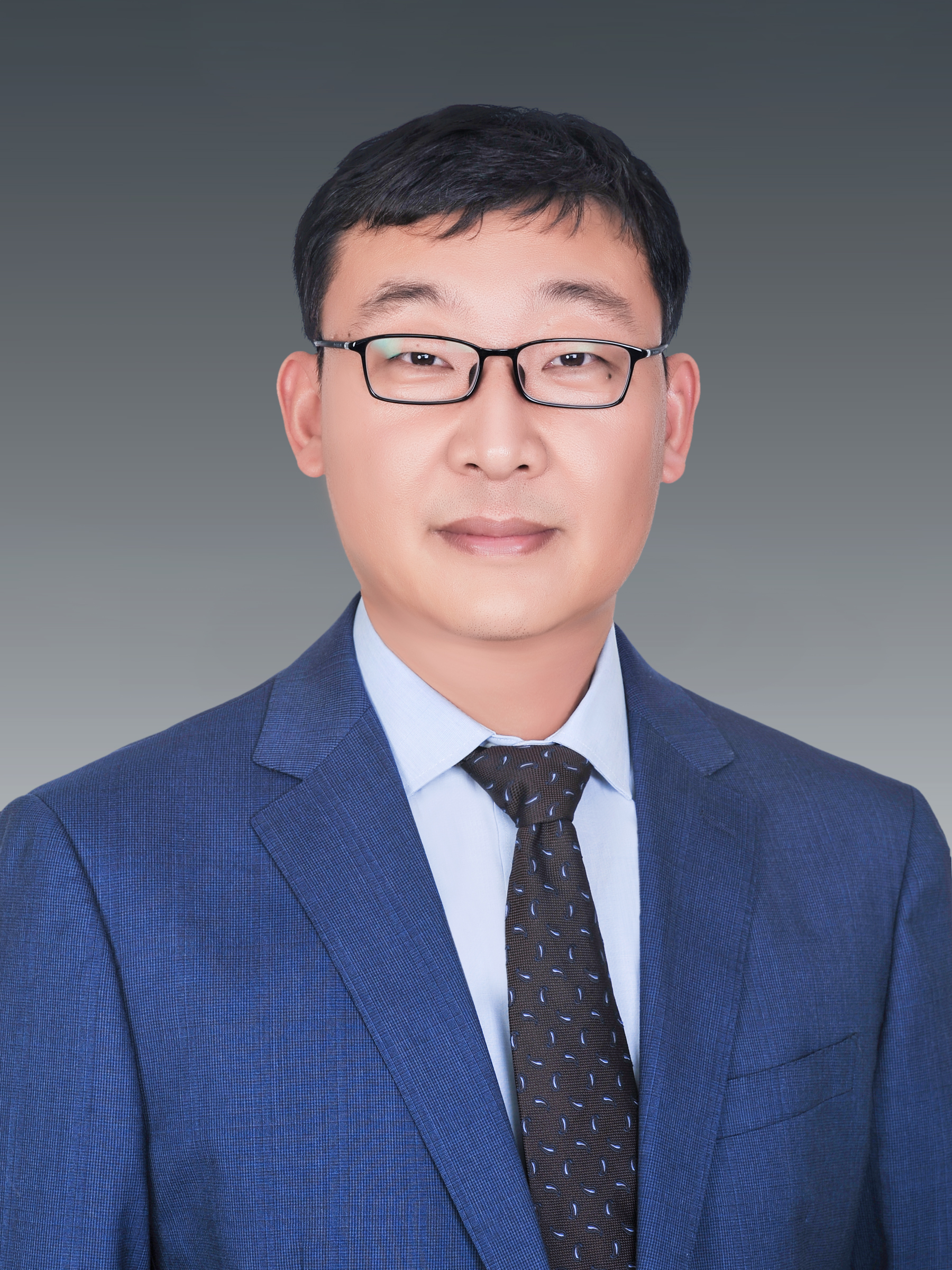}}]{Xin Wang} is an Associate Professor in the School of Artificial Intelligence at Jilin University, China. His research interests include trustworthy graph neural networks, out-of-distribution generalization of graph data, knowledge editing, and fairness in large language models. Dr. Wang is a CCF Outstanding Member and a distinguished talent in Jilin Province. He serves as an executive committee member of the CCF Computer Applications Technical Committee, as well as a member of the Big Data and Artificial Intelligence and Pattern Recognition Technical Committees. He is also a member of the Social Media Processing Technical Committee of the Chinese Information Society. Dr. Wang has led multiple national projects, including NSFC funding, and has published over 20 papers in top AI conferences and journals such as ICML, ICLR, KDD, and ACL. He serves as Area Chair for ACL and Senior Program Committee Member for AAAI.
\end{IEEEbiography}

\end{document}

%% file: 1_intro/intro_v3.tex
Graph Neural Networks (GNNs) have made remarkable advancements in modeling and learning from graph-structured data across various scenarios, including, encompassing social networks~\cite{fan2020graph,chang2021sequential,li2022cyclic}, molecules~\cite{shui2020heterogeneous,liu2024data}, and knowledge graphs~\cite{kg_1,kg_2}. Despite the powerful representational capabilities of GNNs, their success often relies on the assumption that the training and testing data follow the same distribution. Unfortunately, such an assumption rarely holds in most real-world applications, where out-of-distribution (OOD) data from different distributions often occurs~\cite{OODsurvey,Obenchmark}. Empirical evidence has shown that GNNs often struggle to maintain performance when tested on OOD data that differ significantly from the training set~\cite{OODbench,li2022ood}. These vulnerabilities underscore the critical need to enhance the OOD generalization capabilities of GNNs, which has become a rapidly growing area of research~\cite{liu2024arc,pan2023prem,fan2023generalizing}.

{Building on the success of \textit{invariance principle} in addressing OOD generalization challenges in image data~\cite{creager2021environment,ye2022ood}, graph invariant learning~(GIL) has been proposed and recently emerged as a prominent solution for tackling its counterpart problem~\cite{li2022learning,fan2022debiasing,miao2022interpretable,wu2022discovering}.} The fundamental assumption underlying GIL is that each graph can be divided into two distinct components, namely \textit{invariant subgraph} and \textit{environment subgraph}. The former exhibits deterministic and solid predictive relationships with the label of the graph, while the latter may show spurious correlations with the labels and can vary significantly in response to distribution shifts~\cite{yang2022learning,chen2024does}. 
Theoretically, if invariant and environmental information can be accurately separated, GIL-based methods can effectively learn invariant representations from the input graph and make reliable predictions in OOD scenarios. Consequently, existing methods primarily focus on \textbf{capturing and modeling the environmental subgraphs,} employing carefully designed loss functions to minimize the correlations between the predicted environmental subgraphs and the labels~\cite{gui2024joint,piao2024improving}, or utilizing data augmentation strategies to simulate potential environments in wild data~\cite{wu2022handling,jia2024graph}.

\begin{figure}[t!]   

 \centering
 \subfigure[Diversity of environments.]{
   \includegraphics[width=0.21\textwidth]{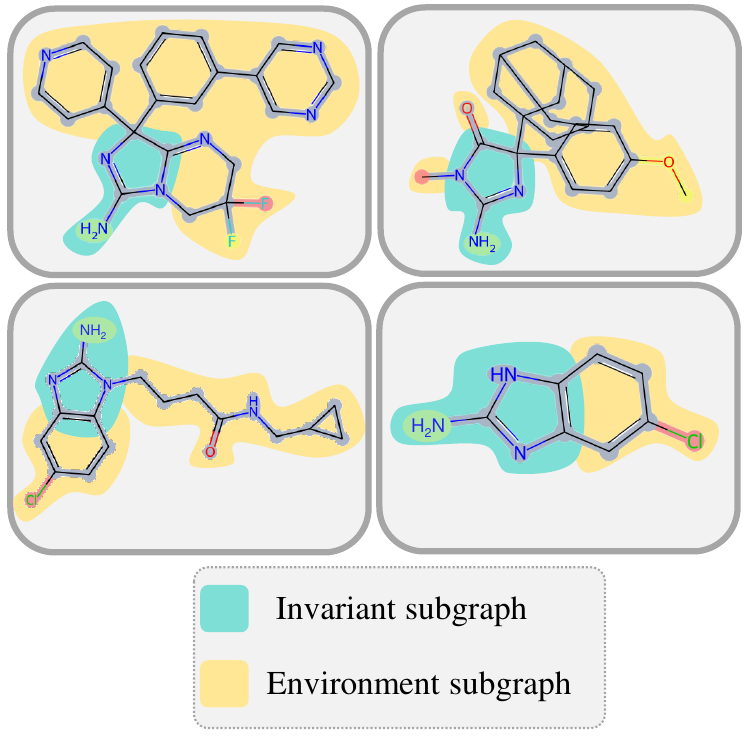}
   \label{subfig:intro1}
 } 
 \subfigure[Semantic cliff across classes.]{
   \includegraphics[width=0.21\textwidth]{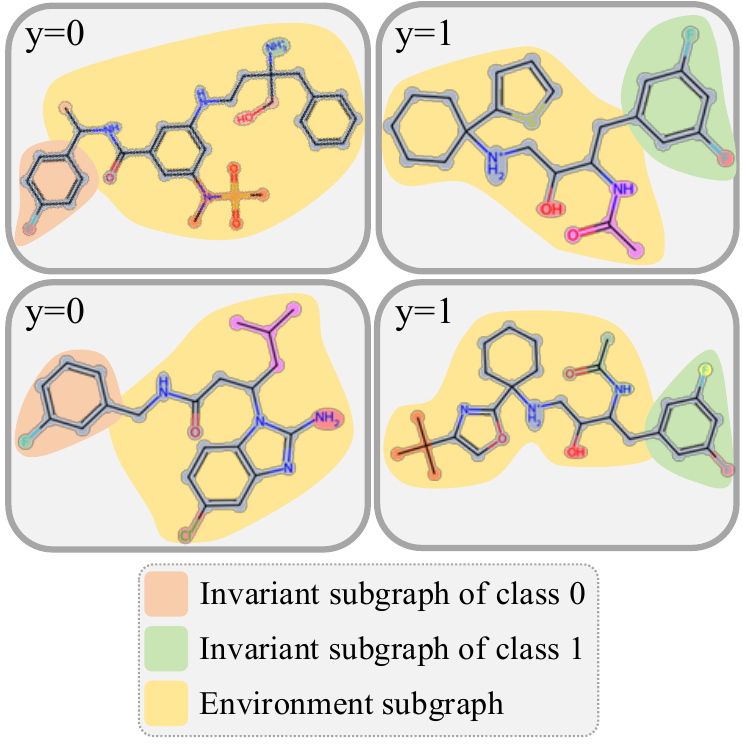}
   \label{subfig:intro2}
}

\caption{Case examples of two challenges faced by graph out-of-dirstribution generalization}

\label{fig:moti}
\end{figure}
{Although GIL-based methods are theoretically viable, they directly apply the concept of environment to the graph domain,  overlooking the \textbf{difficulties in capturing environment information} in real-world graph data.} These difficulties can be attributed to the \textit{diversity}, \textit{distinguishability}, and \textit{lack of labels} of practical graph environments. 
Specifically, the environments of graph data can exhibit substantial diversity in terms of sizes, shapes, and topological properties. Taking molecular graphs as an example, as shown in Fig.~\ref{subfig:intro1}, invariance can be associated with functional groups that determine specific chemical properties, whereas environmental factors manifest as structural variations such as different scaffolds and side chains~\cite{zhuang2023learning}, or assay and size of the molecular~\cite{drugood}. In this context, even with augmented or reorganized environments during training, GNN models struggle to identify all forms of environment subgraphs in real-world OOD samples.  Moreover, unlike image data, where environments can be explicitly defined (e.g., background or style)~\cite{lin2022zin}, the structural boundaries between invariant and environmental subgraphs in graph data are often indistinct, making environmental information difficult to delineate and label. Given the above difficulties, most existing GIL-based approaches struggle to effectively identify environmental information, leading to suboptimal performance, as empirically demonstrated by~\cite{chen2024does}. Consequently, a natural question arises: \textbf{\textit{(RQ1) Can we consider a more feasible way to learn invariant representations on graphs without explicitly modeling the environment information?}}

{Furthermore, recent theoretical studies in OOD generalization emphasize that ensuring \textit{inter-class separation} is essential alongside learning invariant representation~\cite{ye2021towards,bai2024hypo}. However, no studies have yet incorporated this into GIL for graph OOD generalization. Motivated by the inherent properties of graphs~\cite{xia2024understanding}, we introduce another critical challenge in GIL, referred to as the \textbf{semantic cliff across different classes}.}
For instance, as shown in Fig.~\ref{subfig:intro2}, invariant subgraphs (e.g., functional groups) across different molecular classes often share similar structural characteristics, with distinctions frequently limited to a single atom or bond. In such cases, the decision boundaries between invariant representations can become blurred, exacerbating the challenge of separating classes, particularly when the invariant and environment subgraphs are not distinctly identifiable. Nevertheless, existing GIL-based methods primarily emphasize the extraction of invariant information while overlooking their inter-class separability, thereby resulting in suboptimal generalization performance. Given this shortage, a natural follow-up question arises: \textbf{\textit{(RQ2) How can we develop a more robust OOD generalization approach that better discriminates between invariant representations across different classes?}}

In this paper, our unique motivation lies in simultaneously addressing the above two research questions, for which we propose a novel \textbf{M}ulti-\textbf{P}rototype \textbf{H}yperspherical \textbf{I}nvariant \textbf{L}earning (\ourmethod for short) method. \ourmethod is built upon a novel framework, termed hyperspherical graph invariant learning, which incorporates two advanced design: 1)~\textbf{hyperspherical invariant representation extraction}, which maps the extracted invariant representations to hyperspherical space to enhance separability, and 2)~\textbf{multi-prototype hyperspherical classification }, which assigns multiple prototypes to each class and employs them as intermediate variables between the input and labels, thereby eliminating the necessity for explicit environmental modeling.
More specifically, we propose a more practical learning objective for \ourmethod that incorporates two well-crafted loss terms. To address \textbf{\textit{(RQ1)}}, we design an \textit{invariant prototype matching loss} ($\mathcal{L}_{IPM}$) that ensures samples from the same class are always 
assigned to the same class prototype in hyperspherical space. In this way, $\mathcal{L}_{IPM}$ can allow the model to extract robust invariant features across varying environments without explicitly modeling them. To answer \textbf{\textit{(RQ2)}}, we produce a \textit{prototype separation loss} ($\mathcal{L}_{PS}$) that pulls the prototypes belonging to the same class closer together while ensuring those from different classes remain dissimilar. In this way, $\mathcal{L}_{PS}$ enhances the class separability in the context of OOD generalization and mitigates the semantic cliff issue in graph data. 
To sum up, the main contributions of this paper are as follows:

\begin{itemize}[leftmargin=*]
    \item \textbf{Framework.} Derived from the objective of GIL, we introduce a novel hyperspherical graph invariant learning framework along with its corresponding  learning objective, ensuring robust invariant representation learning while eliminating the need for explicit environment modeling. 
    
    \item \textbf{Methodology.} Based on the proposed framework, we develop a novel graph OOD generalization method termed \ourmethod. It incorporates two key innovations: hyperspherical invariant representation extraction and  multi-prototype classification mechanism.
    
    \item \textbf{Experiments.}  We conduct extensive experiments to validate the effectiveness of \ourmethod, and the results demonstrate its superior generalization ability compared to state-of-the-art methods across various types of distribution shifts.
\end{itemize}



%% file: 6_appendix/rw.tex

\subsection{OOD Generalization and Invariant Learning}Due to the sensitivity of deep neural networks to distributional shifts, their performance can vary dramatically, making out-of-distribution (OOD) generalization an important research topic~\cite{arjovsky2019invariant,li2022out,krueger2021out}. The predominant approach for OOD generalization is invariant learning, which explores stable relationships between features and labels across environments, aiming to learn representations that remain effective in OOD scenarios~\cite{creager2021environment,ahuja2020empirical}. Its interpretability is ensured by a causal data generation process~\cite{sun2020latent}. \cite{ye2021towards} proved the theoretical error lower bound for OOD generalization based on invariant learning. Notably, the definition of environmental information is critical to invariant learning, yet it also restricts its further development, as it often requires the training set to encompass a diverse and comprehensive range of environments~\cite{lin2022zin,rosenfeld2020risks,nagarajan2020understanding}.

\subsection{OOD Generalization on Graphs}
Inspired by the sucess of invariant learning in image data, many OOD generalization methods in the graph domain have been proposed, adopting this core idea, and several representative works have emerged.~\cite{yang2022learning,li2022learning,liu2022graph,jia2024graph,fan2022debiasing,sui2022causal,miao2022interpretable}. 
Their core idea is to design an effective model or learning strategy that can identify meaningful invariant subgraphs from the input while ignoring the influence of environmental noise~\cite{wu2022discovering,chen2022learning}. However, the difficulty of modeling environment information in the graph domain has recently garnered attention, with researchers generally agreeing that directly applying invariant learning to graph OOD generalization presents challenges~\cite{chen2024does,zhuang2023learning}.

\subsection{Hyperspherical Learning} Hyperspherical learning has gained attention due to its advantages over traditional Euclidean methods in high-dimensional~\cite{davidson2018hyperspherical,ke2022hyperspherical}. The core idea lies in using a projector to project representations onto a unit sphere space for prototype-based classification~\cite{mettes2019hyperspherical}. Recently, hyperspherical learning has been extended to applications like contrastive learning and OOD detection,  allowing for better disentanglement of features~\cite{ming2022exploit,bai2024hypo}. Despite these advancements, the effective integration of hyperspherical representations with invariant learning to tackle graph OOD generalization has yet to be explored, with the primary challenge being the difficulty of accurately defining environmental information in these tasks.

%% file: 2_preliminary/pre_v2.tex
In this section, we introduce the preliminaries and background of this work, including the formulation of the graph OOD generalization problem, graph invariant learning, and hyperspherical embeddings.

\subsection{Problem Formulation} 
In this paper, we focus on the OOD generalization problem on graph classification tasks~\cite{li2022out,jia2024graph,fan2022debiasing,wu2022discovering}. We denote a graph data sample as $(G,y)$, where $G \in \mathcal{G}$ represents a graph instance and $y \in \mathcal{Y}$ represents its label. The dataset collected from a set of environments $\mathcal{E}$ is denoted as $\mathcal{D} = \{\mathcal{D}^{e}\}_{e \in \mathcal{E}}$, where $\mathcal{D}^{e} =\{({G}^{e}_{i},{y}^{e}_{i})\}^{n^{e}}_{i=1}$ represents the data from environment $e$, and $n^e$ is the number of instances in environment $e$. Each pair $({G}^{e}_{i}, {y}^{e}_{i})$ is sampled independently from the joint distribution $P_{e}(\mathcal{G}, \mathcal{Y}) = P(\mathcal{G}, \mathcal{Y} | e)$. 
In the context of graph OOD generalization, the difficulty arises from the discrepancy between the training data distribution $P_{e_{tr}}(\mathcal{G}, \mathcal{Y})$ from environments $e_{tr} \in \mathcal{E}_{tr}$, and the testing data distribution $P_{e_{te}}(\mathcal{G}, \mathcal{Y})$ from unseen environments $e_{te} \in \mathcal{E}_{test}$, where $\mathcal{E}_{te} \neq \mathcal{E}_{tr}$. The goal of OOD generalization is to learn an optimal predictor $f: \mathcal{G} \rightarrow \mathcal{Y}$ that performs well across both training and unseen environments, $\mathcal{E}_{all} = \mathcal{E}_{tr} \cup \mathcal{E}_{te}$, i.e., 
\begin{equation}
\label{eq: OOD_target} \min_{f \in \mathcal{F}} \max_{e \in \mathcal{E}_{\mathrm{all}}} \mathbb{E}_{(G^{e}, y^{e}) \sim P_{e}}[\ell(f(G^e), y^e)], 
\end{equation}
where $\mathcal{F}$ denotes the hypothesis space, and $\ell(\cdot,\cdot)$ represents the empirical risk function. 

\subsection{Graph Invariant Learning (GIL)}
Invariant learning focuses on capturing representations that preserve consistency across different environments, ensuring that the learned invariant representation $\mathbf{z}_{inv}$ maintains consistency with the label $y$~\cite{mitrovic2020representation,wu2022discovering,chen2022learning}. Specifically, for graph OOD generalization, the objective of GIL is to learn an invariant GNN $f:= f_{c} \circ g$, where $g: \mathcal{G} \rightarrow \mathcal{Z}_{inv}$ is an encoder that extracts the invariant representation from the input graph $G$, and $f_{c}: \mathcal{Z}_{inv} \rightarrow \mathcal{Y}$ is a classifier that predicts the label $y$ based on $\mathbf{z}_{inv}$. From this perspective, the optimization objective of OOD generalization, as stated in Eq.~(\ref{eq: OOD_target}), can be reformulated as: 
\begin{equation}
\label{eq: causal} \max_{f_{c}, g} I(\mathbf{z}_{inv}; y), \text{ s.t. } \mathbf{z}_{inv} \perp e,\forall e \in \mathcal{E}_{tr}, \mathbf{z}_{inv} = g(G),
\end{equation}
{where $I(\mathbf{z}_{inv}; y)$ denotes the mutual information between the invariant representation $\mathbf{z}_{inv}$ and the label $y$.}
This objective ensures that $\mathbf{z}_{inv}$ is independent of the environment $e$, focusing solely on the most relevant information for predicting $y$. 

\subsection{Hyperspherical Embedding} 
Hyperspherical learning enhances the discriminative ability and generalization of deep learning models by mapping feature vectors onto a unit sphere~\cite{liu2017deep}. 
To learn a hyperspherical embedding for the input sample, its representation vector $\mathbf{z}$ is mapped into hyperspherical space with arbitrary linear or non-linear projection functions, followed by normalization to ensure that the projected vector $\hat{\mathbf{z}}$ lies on the unit hypersphere ($\|\hat{\mathbf{z}}\|^{2}=1$). 
To make classification prediction, the hyperspherical embeddings $\hat{\mathbf{z}}$ are modeled using the von Mises-Fisher (vMF) distribution~\cite{ming2022exploit}, with the probability density for a unit vector in class $c$ is given by:
\begin{equation}
\label{eq: vMF} p(\hat{\mathbf{z}}; \boldsymbol{\mu}^{(c)}, \kappa ) = Z(\kappa) \exp(\kappa {\boldsymbol{\mu}^{(c)}}^\top \hat{\mathbf{z}}), 
\end{equation} 
where $\boldsymbol{\mu}^{(c)}$ denotes the prototype vector of class $c$ with the unit norm, serving as the mean direction for class $c$, while $\kappa$ controls the concentration of samples around $\boldsymbol{\mu}_c$.
The term $Z(\kappa)$ serves as the normalization factor for the distribution. Given the probability model in Eq.(\ref{eq: vMF}), the hyperspherical embedding $\hat{\mathbf{z}}$ is assigned to class $c$ with the following probability:
\begin{equation} \label{eq: prototpye_1}
    \begin{aligned}
\mathbb{P}\left(y = c \mid \hat{\mathbf{z}}; \{\kappa, \boldsymbol{\mu}^{(i)}\}_{i = 1}^{C}\right) &= \frac{Z(\kappa) \exp \left(\kappa {\boldsymbol{\mu}^{(c)}}^{\top} \hat{\mathbf{z}}\right)}{\sum_{i = 1}^{C} Z(\kappa) \exp \left(\kappa {\boldsymbol{\mu}^{(i)}}^{\top} \hat{\mathbf{z}}\right)}\\ &= \frac{\exp \left({\boldsymbol{\mu}^{(c)}}^{\top} \hat{\mathbf{z}} / \tau\right)}{\sum_{i = 1}^{C} \exp \left({\boldsymbol{\mu}^{(i)}}^{\top} \hat{\mathbf{z}}/ \tau\right)},
    \end{aligned}
\end{equation}
where $\tau = 1/\kappa$ is a temperature parameter. In this way, the classification problem is transferred to the distance measurement between the graph embedding and the prototype of each class in hyperspherical space, where the class prototype is usually defined as the embedding centroid of each class.

%% file: 3_method/method_v2.tex
\begin{figure*}[t!] 
\centering    
\includegraphics[scale=0.5]{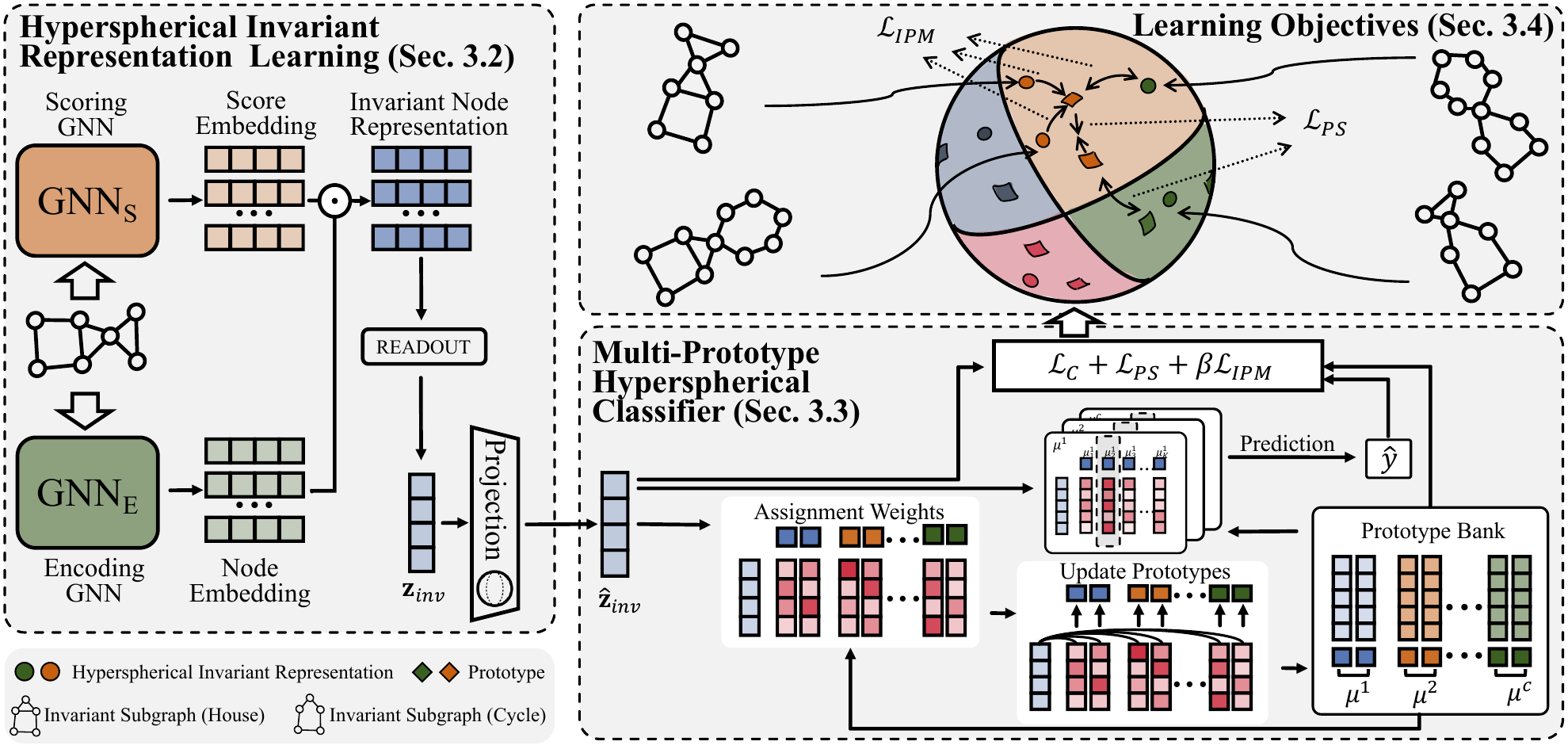}
\caption{The overall framework of \ourmethod. First, a GNN-based model generates the invariant representation and maps it into the hyperspherical space. Then, the classifier makes the prediction based on multiple prototypes. The overall method is trained by a three-term joint objective.}
\label{fig:framework} 
\end{figure*}

In this section,we present the proposed method,\textbf{M}ulti-\textbf{P}rototype
 \textbf{H}yperspherical \textbf{I}nvariant \textbf{L}earning(\ourmethod). In Sec.~\ref{subsec:framework}, we first derive our general framework based on the learning objective graph invariant learning (GIL). Then, we describe the specific designs of the components in MPHIL, including hyperspherical invariant representation learning (Sec.~\ref{subsec:invariant_model}), multi-prototype classifier (Sec.~\ref{subsec:proto}), and learning objectives (Sec.~\ref{subsec:objective}). The overall learning pipeline of \ourmethod is shown in Fig.~\ref{fig:framework}.
\subsection{Hyperspherical Graph Invariant Learning Framework} \label{subsec:framework}
The objective of GIL (i.e., Eq.~(\ref{eq: causal})) aims to maximize the mutual information between the invariant representation $\mathbf{z}_{inv}$ and the label $y$, while ensuring that $\mathbf{z}_{inv}$ remains independent of the environment $e$. However, directly optimizing this objective with such strict constraints is challenging due to the difficulty of modeling environments in graph data. To make the optimization more tractable, we relax the independence constraint and introduce a soft-constrained formulation:
\begin{equation}
\label{eq:soft}
   \underset{f_{c}, g}{\min} -I(y; \mathbf{z}_{inv})+\beta I(\mathbf{z}_{inv};e),
\end{equation}
where $e$ represents the environment to which the current graph belongs, but it cannot be directly observed or accessed. The parameter $\beta$ controls the trade-off between the predictive power of $\mathbf{z}_{inv}$ and its independence from the environment $e$. 

Although the relaxed objective Eq.~(\ref{eq:soft}) is more feasible, the intractable properties of real-world graph data (i.e., \textit{complex environment information} and \textit{inter-class semantic cliff} as discussed in Sec.~\ref{sec:intro}) still hinder us from learning reliable invariant representations and making accurate predictions with this objective. Specifically, the diversity and complexity of environments make it challenging to explicitly model $e$, leading to the difficulties of minimizing $I(\mathbf{z}_{inv};e)$. On the other hand, the semantic cliff issue may cause indistinguishable $\mathbf{z}_{inv}$ of samples belonging to different classes, resulting in the hardness of maximizing $I(y; \mathbf{z}_{inv})$ using a simple cross-entropy loss. 

To deal with the above challenges, we propose a new GIL framework based on hyperspherical space. Concretely, we model the invariant representations in a \textit{hyperspherical space} rather than an arbitrary latent representation space, which enhances the discriminative ability of the learned representations and improves robustness to environmental variations. The desirable properties of hyperspherical space allow us to introduce an intermediate variable, the \textit{class prototype} $\boldsymbol{\mu}$, as a bridge between the hyperspherical invariant representation $\mathbf{z}_{inv}$ and label $y$, which solve the above issues. To be more specific, the class prototypes $\boldsymbol{\mu}$, serving as cluster centers in the class space, directly capture the invariant patterns of each class. By ensuring correct matching between input samples and their corresponding class prototypes, reliable invariant representations can be learned within hyperspherical space, obviating the need for explicit environmental modeling $e$. Moreover, the prototype-based classifier is more robust against the semantic cliff issue, since the semantic gaps between classes can be precisely represented by the relative distances between prototypes in the hyperspherical space. Formally, the reformed learning objective is as follows: 
\begin{equation}\label{eq: target2}
   \underset{f_{c}, g}{\min} \underbrace{-I(y;\hat{\mathbf{z}}_{inv},\boldsymbol{\mu}^{(y)}) }_{\mathcal{L}_{\mathrm{C}}}\underbrace{-I(y ; \boldsymbol{\mu}^{(y)}) }_{\mathcal{L}_{\mathrm{PS}}}\underbrace{-\beta I(\hat{\mathbf{z}}_{inv};\boldsymbol{\mu}^{(y)})}_{\mathcal{L}_{\mathrm{IPM}}},
\end{equation}
where $\hat{\mathbf{z}}_{inv}$ represents the invariant representation in the hyperspherical space and $\boldsymbol{\mu}^{(y)}$ is the prototype corresponding to class $y$. In the following subsections, we first provide the detailed deductions from Eq.~(\ref{eq:soft}) to Eq.~(\ref{eq: target2}) (Sec.~\ref{deduct}), then we will introduce \ourmethod as a practical implementation of the above framework, including the encoder $f_c$ for representation learning (Sec.~\ref{subsec:invariant_model}), the multi-prototype classifier $g$ (Sec.~\ref{subsec:proto}), and the three learning objective terms (Sec.~\ref{subsec:objective}).
\subsection{Proof of the overall objective}\label{deduct}
In this section, we explain how we derived our goal in Eq.~(\ref{eq: target2}) from Eq.~(\ref{eq:soft}). Let's recall that Eq.~(\ref{eq:soft}) is formulated as:
\begin{equation}
   \underset{f_{c}, g}{\min} -I(y; \mathbf{z}_{inv})+\beta I(\mathbf{z}_{inv};e),
\end{equation}
For the first term $-I(y;\mathbf{z}_{inv})$, since we are mapping invariant features to hyperspherical space, we replace $\mathbf{z}_{inv}$ with $\mathbf{\hat{z}}_{inv}$. Then according to the definition of mutual information:
\begin{equation}\label{eq:muinfor}
-I(y;\mathbf{\hat{z}}_{inv}) = - \mathbb{E}_{y, \mathbf{\hat{z}}_{inv}} \left[ \log \frac{p(y,\mathbf{\hat{z}}_{inv})}{p(y) p(\mathbf{\hat{z}}_{inv})} \right].
\end{equation}

We introduce intermediate variables $\boldsymbol{\mu}^{y}$ to rewrite Eq. (\ref{eq:muinfor}) as:
\begin{equation}
    \resizebox{.9\hsize}{!}{
$-I(y; \mathbf{\hat{z}}_{inv}) = -E_{y, \mathbf{\hat{z}}_{inv}, \boldsymbol{\mu}^{y}} \left[ \log \frac{p(y, \mathbf{\hat{z}}_{inv}, \boldsymbol{\mu}^{y})}{p(\mathbf{\hat{z}}_{inv}, \boldsymbol{\mu}^{y}) p(y)} \right] + E_{y, \mathbf{\hat{z}}_{inv},\boldsymbol{\mu}^{y}} \left[ \log \frac{p(y,\boldsymbol{\mu}^{y}|\mathbf{\hat{z}}_{inv})}{p(y|\mathbf{\hat{z}}_{inv}) p(\boldsymbol{\mu}^{y})|\mathbf{\hat{z}}_{inv}} \right]. $}
\end{equation}

By the definition of Conditional mutual information, we have the following equation:
\begin{equation}
    \begin{aligned} 
    -I(y;\mathbf{\hat{z}}_{inv})  &= -I(y;\mathbf{\hat{z}}_{inv},\boldsymbol{\mu}^{y})+I(y;\boldsymbol{\mu}^{y}|\mathbf{\hat{z}}_{inv}),  \\ 
    -I(y;\boldsymbol{\mu}^{y})  &= -I(y;\mathbf{\hat{z}}_{inv},\boldsymbol{\mu}^{y})+I(y;\mathbf{\hat{z}}_{inv}|\boldsymbol{\mu}^{y}). 
\end{aligned}
\end{equation}

By merging the same terms, we have:
\begin{equation}
     -I(y;\mathbf{\hat{z}}_{inv})  = -I(y;\boldsymbol{\mu}^{y})+[I(y;\boldsymbol{\mu}^{y}|\mathbf{\hat{z}}_{inv})-I(y;\mathbf{\hat{z}}_{inv}|\boldsymbol{\mu}^{y})]. 
\end{equation}

Since our classification is based on the distance between $\boldsymbol{\mu}^{y}$ and $\mathbf{\hat{z}}_{inv}$, we add $-I(y;\mathbf{\hat{z}}_{inv},\boldsymbol{\mu}^{y})$ back into the above equation and obtain a lower bound:
\begin{equation}
\resizebox{.9\hsize}{!}{
$-I(y;\mathbf{\hat{z}}_{inv}) \geq -I(y;\boldsymbol{\mu}^{y})+[I(y;\boldsymbol{\mu}^{y}|\mathbf{\hat{z}}_{inv})-I(y;\mathbf{\hat{z}}_{inv}|\boldsymbol{\mu}^{y})]-I(y;\mathbf{\hat{z}}_{inv},\boldsymbol{\mu}^{y}).$}
\end{equation}
Since the $\boldsymbol{\mu}^{y}$ are updated by $\mathbf{\hat{z}}_{inv}$ from the same class, we can approximate $I(y;\boldsymbol{\mu}^{y}|\mathbf{\hat{z}}_{inv})$ equal to $I(y;\mathbf{\hat{z}}_{inv}|\boldsymbol{\mu}^{y})$ and obtain the new lower bound:
\begin{equation}
\label{eq: first part}
    -I(y;\mathbf{\hat{z}}_{inv}) \geq -I(y;\boldsymbol{\mu}^{y})-I(y;\mathbf{\hat{z}}_{inv},\boldsymbol{\mu}^{y}).
\end{equation}

For the second term $I(\mathbf{z}_{inv};e)$, we can also rewrite it as:
\begin{equation}
  \begin{aligned}
    I(\mathbf{\hat{z}}_{inv};e)  &= I(\mathbf{\hat{z}}_{inv};e,\boldsymbol{\mu}^{y})-I(\mathbf{\hat{z}}_{inv};\boldsymbol{\mu}^{y}|e).
    \end{aligned}
\end{equation}

Given that the environmental labels  $e$ are unknown, we drop the term $I(\mathbf{\hat{z}}_{inv};e,\boldsymbol{\mu}^{y})$ as it cannot be directly computed. This leads to the following lower bound:
\begin{equation}\label{z_e}
  \begin{aligned}
    I(\mathbf{\hat{z}}_{inv};e)  &\geq-I(\mathbf{\hat{z}}_{inv};\boldsymbol{\mu}^{y}|e).
    \end{aligned}
\end{equation}

We can obtain a achievable target by Eq. (\ref{eq: first part}) and Eq. (\ref{z_e}) as follow:
\begin{equation}\label{uneq_1}
\resizebox{.9\hsize}{!}{$
    -I(y; \mathbf{z}_{inv})+\beta I(\mathbf{z}_{inv};e)\geq-I(y;\boldsymbol{\mu}^{y})-I(y;\mathbf{\hat{z}}_{inv},\boldsymbol{\mu}^{y})-\beta I(\mathbf{\hat{z}}_{inv};\boldsymbol{\mu}^{y}|e).$}
\end{equation}

In fact, $p(\mathbf{\hat{z}}_{inv},\boldsymbol{\mu}^{y}|e) \geq p(\mathbf{\hat{z}}_{inv};\boldsymbol{\mu}^{y})$, Eq. (\ref{uneq_1}) can be achieved by:
\begin{equation}\resizebox{.9\hsize}{!}{$
    -I(y; \mathbf{z}_{inv})+\beta I(\mathbf{z}_{inv};e)\geq-I(y;\boldsymbol{\mu}^{y})-I(y;\mathbf{\hat{z}}_{inv},\boldsymbol{\mu}^{y})-\beta I(\mathbf{\hat{z}}_{inv};\boldsymbol{\mu}^{y}). $}
\end{equation}

Finally, optimizing Eq. (\ref{eq:soft}) can be equivalent to optimizing its lower bound and we can obtain the objective without environment $e$ as shown in Eq. (\ref{eq: target2}):
\begin{equation}
\setlength\abovedisplayskip{9pt}
\setlength\belowdisplayskip{9pt}
   \underset{f_{c}, g}{\min} \underbrace{-I(y;\hat{\mathbf{z}}_{inv},\boldsymbol{\mu}^{(y)}) }_{\mathcal{L}_{\mathrm{C}}}\underbrace{-I(y ; \boldsymbol{\mu}^{(y)}) }_{\mathcal{L}_{\mathrm{PS}}}\underbrace{-\beta I(\hat{\mathbf{z}}_{inv};\boldsymbol{\mu}^{(y)})}_{\mathcal{L}_{\mathrm{IPM}}}.
\end{equation}
\subsection{Hyperspherical Invariant Representation Extraction} \label{subsec:invariant_model}

\noindent\textbf{Encoder.} 
In GIL, the goal of the encoder $f$ is to extract invariant representations that are highly correlated with the invariant subgraph of each sample. Nevertheless, explicitly identifying the subgraphs via modeling the selecting probabilities of each node and edge may lead to increased overhead and require more complex network architectures~\cite{zhuang2023learning}. To mitigate these costs, we adopt a lightweight GNN-based model for efficient invariant representation learning. To be specific, our model includes two GNNs: $\mathrm{GNN}_{E}$ to encode the input graph $G$ into the latent space, producing the node representation $\mathbf{H}$, and $\mathrm{GNN}_{S}$ to compute the separation score $S$ for the invariant features: 
\begin{equation}\label{eq:score}
    \mathbf{H} = \mathrm{GNN}_{E}(G) \in \mathbb{R}^{|\mathcal{V}| \times d}, \mathbf{S} = \sigma(\mathrm{GNN}_{S}(G)) \in \mathbb{R}^{|\mathcal{V}| \times d},
\end{equation}
\noindent where $|\mathcal{V}|$ is the number of nodes in the graph $G$, $d$ is the latent dimension, and $\sigma(\cdot)$ is the Sigmoid function to constrain $\mathbf{S}$ falls into the range of $(0,1)$. Then, the invariant representation $\mathbf{z}_{inv}$ is obtained through the following operation:
\begin{equation}\label{eq:readout}
    \mathbf{z}_{inv} = \mathrm{READOUT}(\mathbf{H} \odot \mathbf{S}) \in \mathbb{R}^{d},
\end{equation}
where $\odot$ is the element-wise product and $\mathrm{READOUT}(\cdot)$ is an aggregation function (e.g., $\mathrm{mean}$) to generate a graph-level representation. 

\noindent\textbf{Hyperspherical Projection.} 
After obtaining $\mathbf{z}_{inv}$, the key step in \ourmethod is to project it into hyperspherical space. Concretely, the hyperspherical invariant representation $\hat{\mathbf{z}}_{inv}$ can be calculated by: 
\begin{equation}
\label{eq: hype_rinv}
    \tilde{\mathbf{z}}_{inv} = \mathrm{Proj}(\mathbf{z}_{inv}), \hat{\mathbf{z}}_{inv} = \tilde{\mathbf{z}}_{inv} / \| \tilde{\mathbf{z}}_{inv}\|_{2}, 
\end{equation}
where $\mathrm{Proj}(\cdot)$ is a MLP-based projector that maps the representation into another space, and dividing $\tilde{\mathbf{z}}_{inv}$ by its norm constrains the representation vector to unit length. The hyperspherical projection allows invariant learning to occur in a more discriminative space. More importantly, the hyperspherical space provides a foundation for prototypical classification, enabling the extraction of invariant patterns without modeling environments and addressing the semantic cliff issue.

\subsection{Multi-Prototype Hyperspherical Classifier} \label{subsec:proto}
Following hyperspherical projection, the next step is to construct a prototype-based classifier within the hyperspherical space. In conventional hyperspherical learning approaches~\cite{ke2022hyperspherical}, each class is typically assigned a single prototype. Although this is a straightforward solution, its modeling capabilities regarding decision boundaries are limited, as it may not adequately capture the complexity of the data distribution. More specifically, a single prototype often overfits easy-to-classify samples while failing to consider the harder samples. To address this limitation, we propose a multi-prototype hyperspherical classifier in which \textit{each class is represented by multiple prototypes}. This multi-prototype approach ensures that the classification decision space is more flexible and comprehensive, enabling better modeling of intra-class invariance and inter-class separation. In the following paragraphs, we will explain how to initialize, update, and use the prototypes for prediction.

\noindent\textbf{Prototype Initialization.} 
For each class $c \in \{1, \cdots, C\}$, we assign $K$ prototypes for it, and they can be denoted by $\mathbf{M}^{(c)} \in \mathbb{R}^{K \times d} = \{ \boldsymbol{\mu}_{k}^{(c)}\}^{K}_{k=1}$. At the beginning of model training, we initialize each of them by $\boldsymbol{\mu}_{k}^{(c)} \sim \mathcal{N}(\textbf{0}, \textbf{I})$, where $\mathcal{N}(\textbf{0}, \textbf{I})$ represents a standard multivariate Gaussian distribution. Random initialization can help prevent the issue of mode collapse.

\noindent\textbf{Prototype Updating.} 
To ensure that the prototypes can represent the majority of samples in their corresponding classes while preserving  cross-distribution stability, we adopt the exponential moving average (EMA) technique to update the prototypes asynchronously according to invariant representation $\hat{\mathbf{z}}_{inv}$. Specifically, the update rule for a batch of $B$ samples is given by:
\begin{equation} \label{eq: update_prototype}
    \boldsymbol{\mu}_{k}^{(c)} := \text{Norm}\left( \alpha \boldsymbol{\mu}_{k}^{(c)} + (1 - \alpha) \sum_{i=1}^{B} 1(y_i = c) \mathbf{W_{i,k}^{(c)}} \mathbf{\hat{Z}_{inv,i}} \right), 
\end{equation}
where $\alpha$ is the EMA update rate, $\mathbf{W_{i,k}^{(c)}} $ is the weight of the $i$-th sample for prototype $k$ in class $c$, $\mathbf{\hat{Z}_{inv,i}}$ is the representations of the $i$-th sample, and ${1}(y_i = c)$ is an indicator function that ensures the update applies only to samples of class $c$.  After each update, the prototype is normalized to maintain its unit norm, ensuring it remains on the hypersphere and the distance calculations take place in the same unit space as $\mathbf{\hat{Z}_{inv,i}}$. Each representation $\mathbf{\hat{Z}_{inv,i}}$ is associated with multiple prototypes, weighted by the assignment weight vector $\mathbf{W_{i}^{(c)}} \in \mathbb{R}^{K}$. 

\noindent\textbf{Assignment Weight Calculation.} 
To ensure that each sample is matched with the most relevant prototype, we introduce an attention-based matching mechanism. This approach computes the attention score between each sample and its class prototypes to determine the assignment weights:
\begin{equation}\label{eq: att}
    \mathbf{Q} = \mathbf{\hat{Z}_{inv,i}}\mathbf{W}_{Q}, K = \mathbf{M}^{(c)}\mathbf{W}_{K}, \mathbf{W^{(c)}_i} = \mathrm{softmax}(\frac{\mathbf{Q}\mathbf{K}^\top}{\sqrt{d'}}),
\end{equation}
where $\mathbf{W}_{Q}, \mathbf{W}_{K} \in \mathbb{R}^{d \times {d'}}$ are the learnable weight matrices for the samples and the prototypes, respectively, and $d'$ is the dimension of the projected space. The attention mechanism ensures that the prototype $\boldsymbol{\mu}^{(c)}$ most similar to the current $\mathbf{\hat{Z}_{inv,i}}$ receives the highest weight, which improves classification accuracy and helps the prototype remain aligned with its class center. 


\noindent\textbf{Weight Pruning.}
In practice, to ensure that each sample can only concentrate on a limited number of prototypes: we only preserve $\mathbf{W_{i,k}^{(c)}}$ with the top-$n$ largest values while setting the rest to be 0. 
The motivation is: Directly assigning weights to all prototypes within a class can lead to excessive similarity between prototypes, especially for difficult samples. This could blur decision boundaries and reduce the model's ability to correctly classify hard-to-distinguish samples.

We apply a \textbf{top-$n$ pruning} strategy, which keeps only the most relevant prototypes for each sample. The max weights are retained, and the rest are pruned as follows:
\begin{equation}
    W_{i,k}^{(c)} = {1}[W_{i,k}^{(c)}>\beta]\ast W_{i,k}^{(c)}, 
\end{equation}
where $\beta$ is the threshold corresponding to the top-$n$ weight, and ${1}[W_{i,k}^{(c)}>\beta]$ is an indicator function that retains only the weights for the top-$n$ prototypes. This pruning mechanism ensures that the prototypes remain distinct and that the decision space for each class is well-defined, allowing for improved classification performance. By applying this attention-based weight calculation and top-$n$ pruning, the model ensures a more accurate and robust matching of samples to prototypes, enhancing classification, especially in OOD scenarios.

\noindent\textbf{Prototype-Based Prediction.} 
To make classification decisions with the multi-prototype classifier, we can calculate the prediction probability with the similarity between the invariant representation $\hat{\mathbf{z}}_{inv}$ and the set of prototypes $\boldsymbol{\mu}^{(c)}$ associated with each class, which is defined as:
\begin{align}\label{classify}
\resizebox{.9\hsize}{!}{$
p(y = c | \hat{\mathbf{z}}_{inv}; \{ \mathbf{w}^{(c)}, \boldsymbol{\mu}^{(c)}\}_{c=1}^{(C)}) = 
\frac{\underset {k=1,\dots,K}{\max} w_{k}^{(c)} \exp \left(  {\boldsymbol{\mu}_{k}^{(c)}}^\top \hat{\mathbf{z}}_{inv} / \tau \right)}
{\sum_{j=1}^{C} \underset {{k}=1,\dots,K}{\max} w_{k}^{(j)} \exp \left(  {\boldsymbol{\mu}_{k}^{(j)}}^\top \hat{\mathbf{z}}_{inv} / \tau \right)},$}
\end{align}
{where  $w_{k}^{c}$ represents the weight of the $k$-th prototype $\boldsymbol{\mu}_{k}^{(c)}$ assigned to the current sample for class $c$.}
After that, the class prediction can be directly obtained by an $\mathrm{argmax}$ operation. We classify using the prototype most similar to the sample, as it offers the most representative and discriminative information, helping the sample converge faster to the correct class.

\subsection{\ourmethod Learning Objectives} \label{subsec:objective}

In this subsection, we formulate the learning objective terms of \ourmethod in Eq.~(\ref{eq: target2}), including the invariant prototype matching loss $\mathcal{L}_{\mathrm{IPM}}$, prototype separation loss $\mathcal{L}_{\mathrm{PS}}$, and the classification loss $\mathcal{L}_{\mathrm{C}}$. For $\mathcal{L}_{\mathrm{IPM}}$ and $\mathcal{L}_{\mathrm{PS}}$, we formulate them with contrastive learning loss, which is proved to be an effective mutual information estimator~\cite{sordoni2021decomposed,xie2022self,sun2024interdependence}. For the term of $-I(y;\hat{\mathbf{z}}_{inv},\boldsymbol{\mu}^{(y)})$, we will provide a clear derivation showing that it can be implemented with classification loss.

\noindent\textbf{Invariant Prototype Matching Loss $\mathcal{L}_{\mathrm{IPM}}$.}
The challenge of disentangling invariant features from environmental variations lies at the heart of OOD generalization. In our formulation, the misalignment of a sample with an incorrect prototype can be seen as {a signal of environmental interference.} In contrast, successful alignment with the correct prototype reflects {the capture of stable and invariant features.} Motivated by this, we design $\mathcal{L}_{\mathrm{IPM}}$ that operates by reinforcing the proximity of samples to their invariant representations and penalizing the influence of environmental factors, implicitly captured through incorrect prototype associations.
The loss function is expressed as follows:
\begin{align}\label{ipm}\resizebox{.9\hsize}{!}{$ \mathcal{L}_{\mathrm{IPM}} = - \frac{1}{B} \sum_{i=1}^{B} \log \frac{\sum_{c=y_i} \exp \left( \hat{\mathbf{z}}_i^\top \boldsymbol{\mu}^{(c)} / \tau \right)}{\sum_{c=y_i} \exp \left( \hat{\mathbf{z}}_i^\top \boldsymbol{\mu}^{(c)} / \tau \right) + \sum_{\hat{c} \neq y_i} \exp \left( \hat{\mathbf{z}}_i^\top \boldsymbol{\mu}^{(\hat{c})} / \tau \right)},$}
\end{align}
where $B$ represents the batch size, with $i$ indexing each sample in the batch. $\hat{\mathbf{z}}_i$ represents the hyperspherical invariant representation, $\boldsymbol{\mu}^{(c)}$ is the correct class prototype, $\boldsymbol{\mu}^{(\hat{c})}$ denotes the prototypes of the incorrect classes $\hat{c} \neq y_i$, and $\tau$ is a temperature factor. This formulation reflects the dual objective of pulling samples towards their class-invariant prototypes while ensuring that the influence of prototypes associated with environmental shifts is minimized. The numerator reinforces the similarity between the sample’s invariant representation and its correct prototype, while the denominator introduces competition between correct and incorrect prototypes, implicitly modeling the influence of environmental noise.

\noindent\textbf{Prototype Separation Loss $\mathcal{L}_{\mathrm{PS}}$.} 
In hyperspherical space, all invariant $\hat{\mathbf{z}}$ representations are compactly clustered around their respective class prototypes. To ensure inter-class separability, prototypes of different classes must be distinguishable. The prototype separation loss $\mathcal{L}_{\text{PS}}$ is designed to enforce this by maximizing the separation between prototypes of different classes while encouraging the similarity of prototypes within the same class. The loss function is defined as:
\begin{equation}\label{ps}
\resizebox{.9\hsize}{!}{$
    \mathcal{L}_{\text{PS}} = -\frac{1}{CK} \sum_{c=1}^{C} \sum_{k=1}^{K} \log \frac{\sum_{i = 1}^{K} \mathbb{I}(i \neq k) \exp \left( {(\boldsymbol{\mu}_{k}^{(c)})^\top \boldsymbol{\mu}_{i}^{(c)}}/{\tau} \right)}{\sum_{c' = 1}^{C} \sum_{j = 1}^{K} \mathbb{I}(c' \neq c) \exp \left( {(\boldsymbol{\mu}_{k}^{(c)})^\top \boldsymbol{\mu}_{j}^{(c')}}/{\tau} \right)},$}
\end{equation}
where $C$ represents the total number of classes, $K$ denotes the number of prototypes assigned to each class, $\boldsymbol{\mu}_{k}^{(c)}$ and $\boldsymbol{\mu}_i^{(c)}$ correspond to different prototypes within the same class $c$,
$\boldsymbol{\mu}_{k}^{(c)}$ and $\boldsymbol{\mu}_j^{(c')}$ represent prototypes from different classes, and $\boldsymbol{\mu}_{k}^{(c)}$ and $\boldsymbol{\mu}_j^{(c')}$ represent prototypes from different classes. Such an indicator function ensures that the comparisons are made between distinct prototypes, enhancing intra-class similarity and inter-class separation.

\noindent\textbf{Classification Loss $\mathcal{L}_{\mathrm{C}}$.} 
To obtain the specific implementation of $\mathcal{L}_{\mathrm{C}}$, we first perform the following derivation.
For the term $I(y;\hat{\mathbf{z}}_{inv},\boldsymbol{\mu}^{(y)})$, it can be written as:
\begin{equation}
    I(y;\hat{\mathbf{z}}_{inv},\boldsymbol{\mu}^{(y)}) = E_{y, \mathbf{\hat{z}}_{inv}, \boldsymbol{\mu}^{y}} \left[ \log \frac{p(y| \mathbf{\hat{z}}_{inv}, \boldsymbol{\mu}^{y})}{p(y)} \right],
\end{equation}
according to ~\cite{graphpro}, we have:
\begin{equation}
    I(y;\hat{\mathbf{z}}_{inv},\boldsymbol{\mu}^{(y)}) \geq E_{y, \mathbf{\hat{z}}_{inv}, \boldsymbol{\mu}^{y}} \left[ \log \frac{q_{\theta}(y| \gamma(\mathbf{\hat{z}}_{inv}, \boldsymbol{\mu}^{y}))}{p(y)} \right],
\end{equation}
where $\gamma(,)$ is the function to calculate the similarity between $\mathbf{\hat{z}}_{inv}$ and $\boldsymbol{\mu}^{y}$. $q_{\theta}(y| \gamma(\mathbf{\hat{z}}_{inv}, \boldsymbol{\mu}^{y}))$ is the variational approximation of the $p(y| \gamma(\mathbf{\hat{z}}_{inv}, \boldsymbol{\mu}^{y}))$. Then we can have:
\begin{align}\nonumber
    I(y;\hat{\mathbf{z}}_{inv},\boldsymbol{\mu}^{(y)}) &\geq E_{y, \mathbf{\hat{z}}_{inv}, \boldsymbol{\mu}^{y}} \left[ \log \frac{q_{\theta}(y| \gamma(\mathbf{\hat{z}}_{inv}, \boldsymbol{\mu}^{y}))}{p(y)} \right] \\ \nonumber
    &\geq E_{y, \mathbf{\hat{z}}_{inv}, \boldsymbol{\mu}^{y}} \left[ \log {q_{\theta}(y| \gamma(\mathbf{\hat{z}}_{inv}, \boldsymbol{\mu}^{y}))} \right]\\
    &:= -\mathcal{L}_{\mathrm{C}}.
\end{align}

Finally, we prove that $\min I(y;\hat{\mathbf{z}}_{inv},\boldsymbol{\mu}^{(y)})$ is equivalent to minimizing the classification loss $\mathcal{L}_{\mathrm{C}}$.

To calculate the classification loss with the multi-prototype classifier, we update the classification probability in Eq.~(\ref{classify}) to be closed to truth labels with a classification loss. Take multi-class classification as example, we use the cross-entropy loss:
\begin{equation}\label{cls}
    \mathcal{L}_{\mathrm{C}} = - \frac{1}{BC} \sum_{i=1}^{B}\sum_{c=1}^{C}y_{ic}\text{log}(p(y = c \mid \hat{\mathbf{z}}_{i}; \{ w^{c},\boldsymbol{\mu}^{(c)} \}_{c=1}^{(C)})).
\end{equation}
With the above loss terms, the final objective of \ourmethod can be written as $\mathcal{L}=\mathcal{L}_{\mathrm{C}}+\mathcal{L}_{\mathrm{PS}}+\beta\mathcal{L}_{\mathrm{IPM}}$. 

\subsection{Overall Algorithm of \ourmethod} 
The training algorithm of \ourmethod is shown in Algorithm.~\ref{code:train}. After that, we use the well-trained $\mathrm{GNN}_{S}$,$\mathrm{GNN}_{E}$, $\mathrm{Proj}$ and all prototypes $\mathbf{M}^{(c)} = \{ \boldsymbol{\mu}_{k}^{(c)}\}^{K}_{k=1}$ to perform inference on the test set. The pseudo-code for this process is shown in Algorithm.~\ref{code:test}.
\begin{algorithm}[h!]
\renewcommand{\algorithmicrequire}{\textbf{Input:}}
	\renewcommand{\algorithmicensure}{\textbf{Output:}}
  \caption{The training algorithm of \ourmethod.}
  \begin{algorithmic}[1]
    \REQUIRE
      Scoring GNN $\mathrm{GNN}_{S}$;
      Encoding GNN $\mathrm{GNN}_{E}$;
      Projection $\mathrm{Proj}$;
      Number of prototypes for each class $K$;
      The data loader of in-distribution training set $D_{\mathrm{train}}$.
      
    \ENSURE
    Well-trained $\mathrm{GNN}_{S}$, $\mathrm{GNN}_{E}$, $\mathrm{Proj}$
       and all prototypes $\mathbf{M}^{(c)}$.
    \STATE 
For each class $c \in \{1, \cdots, C\}$, assign $K$ prototypes for it which can be denoted by $\mathbf{M}^{(c)} = \{ \boldsymbol{\mu}_{k}^{(c)}\}^{K}_{k=1}$.    
\STATE Initialize each of them by $\boldsymbol{\mu}_{k}^{(c)} \sim \mathcal{N}(\textbf{0}, \textbf{I})$
     \FOR{epoch in epochs}  
    \FOR{each $G_\mathrm{{batch}}$ in $D_{\mathrm{train}}$}
       \STATE Obtain $Z_{inv}$  using $\mathrm{GNN}_{S}$ and $\mathrm{GNN}_{E}$ via Eq. (\ref{eq:score}) and (\ref{eq:readout})
       \STATE Obtain $\hat{Z}_{inv}$ using $\mathrm{Proj}$ via Eq. (\ref{eq: hype_rinv})
       \STATE Compute $W^{(c)}$ using $\boldsymbol{u}^{(c)}$ and $\hat{Z}_{inv}$ via Eq. (\ref{eq: att}).
        \FOR{each prototype $\boldsymbol{u}_{k}^{(c)}$}
        \STATE Update it using $\hat{Z}_{inv}$ and $W^{(c)}$ via Eq. (\ref{eq: update_prototype}).
        
        \ENDFOR
        \STATE Get $p(y = c \mid \hat{\mathbf{z}}_{i}; \{ w^{c},\boldsymbol{u}^{(c)} \}_{c=1}^{(C)})$ using $\hat{Z}_{inv}$, $W^{(c)}$ and $\boldsymbol{\mu}^{(c)}$ via Eq. (\ref{classify})
        \STATE Compute the final loss $\mathcal{L}$ with $\hat{Z}_{inv}$, $\boldsymbol{\mu}^{(c)}$ and $p(y = c \mid \hat{\mathbf{z}}_{i}; \{ w^{c},\boldsymbol{u}^{(c)} \}_{c=1}^{(C)})$ via Eq. (\ref{ipm}), (\ref{ps}) and (\ref{cls})
        \STATE Update parameters of $\mathrm{GNN}_{S}$, $\mathrm{GNN}_{E}$ and $\mathrm{Proj}$ with the gradient of $\mathcal{L}$.
      \ENDFOR
    \ENDFOR
  \end{algorithmic}\label{code:train}
\end{algorithm}
\vspace{-5mm}
\begin{algorithm}[h!]
\renewcommand{\algorithmicrequire}{\textbf{Input:}}
	\renewcommand{\algorithmicensure}{\textbf{Output:}}
  \caption{The inference algorithm of \ourmethod.}
  \begin{algorithmic}[1]
    \REQUIRE
      Well-trained $\mathrm{GNN}_{S}$,$\mathrm{GNN}_{E}$, $\mathrm{Proj}$ and all prototypes $\mathbf{M}^{(c)} = \{ \boldsymbol{\mu}_{k}^{(c)}\}^{K}_{k=1}$.
      The data loader of Out-of-distribution testing set $D_{\mathrm{test}}$.
      
    \ENSURE
    The final Classification probability $p(y=c \mid \hat{\mathbf{z}}_{i}; \{ w^{c},\boldsymbol{\mu}^{(c)} \}_{c=1}^{(C)})$ 
    \FOR{each $G_\mathrm{{batch}}$ in $D_{\mathrm{test}}$}
       \STATE Obtain $Z_{inv}$  using $\mathrm{GNN}_{S}$ and $\mathrm{GNN}_{E}$ via Eq. (\ref{eq:score}) and (\ref{eq:readout})
       \STATE Obtain $\hat{Z}_{inv}$ using $\mathrm{Proj}$ via Eq. (\ref{eq: hype_rinv})
       \STATE Compute $W^{(c)}$ using $\boldsymbol{\mu}^{(c)}$ and $\hat{Z}_{inv}$ via Eq. (\ref{eq: att}).
        
        \STATE Get $p(y = c \mid \hat{\mathbf{z}}_{i}; \{ w^{c},\boldsymbol{\mu}^{(c)} \}_{c=1}^{(C)})$ using $\hat{Z}_{inv}$, $W^{(c)}$ and $\boldsymbol{\mu}^{(c)}$ via Eq. (\ref{classify})
        
        
      \ENDFOR
  \end{algorithmic}\label{code:test}

\end{algorithm}

\subsection{Complexity Analysis}
The time complexity of \ourmethod is $
\mathcal{O}(|E|d+|V|d^{2})$, where $|V|$ denotes the number of nodes and $|E|$ denotes the number of edges, $d$ is the dimension of the final representation. Specifically, for $\mathrm{GNN}_{S}$ and $\mathrm{GNN}_{E}$, their complexity is denoted as $
\mathcal{O}(|E|d+|V|d^{2})$. The complexity of the projector is $
\mathcal{O}(|V|d^{2})$, while the complexities of calculating weights and updating prototypes are $
\mathcal{O}(|V||K|d)$ where $K$ is the number of prototypes. The complexity of computing the final classification probability also is $
\mathcal{O}(|V|Kd)$. Since $K$ is a very small constant, we can ignore $\mathcal{O}(|V|Kd)$, resulting in a final complexity of $
\mathcal{O}(|E|d+|V|d^{2})$. Theoretically, the time complexity of \ourmethod is on par with the existing methods.

%% file: 4_exp/exp_v2.tex
In this section, we present our experimental setup  (Sec.~\ref{subsec:setup}) and showcase the results in (Sec.~\ref{subsec:results}). For each experiment, we first highlight the research question being addressed, followed by a detailed discussion of the findings.

\begin{table}[t]
\renewcommand{\arraystretch}{1.5}
\setlength\tabcolsep{3pt}
\centering
\caption{Dateset statistics.}
\resizebox{0.49\textwidth}{!}{%
\begin{tabular}{cccccccc}
\toprule
\multicolumn{3}{c|}{Dataset}                                                                            & Task                       & Metric  & Train & Val   & Test  \\ \midrule
\multicolumn{1}{c|}{\multirow{7}{*}{GOOD}}    & \multirow{2}{*}{HIV}   & \multicolumn{1}{c|}{scaffold} & Binary Classification      & ROC-AUC & 24682 & 4133  & 4108  \\
\multicolumn{1}{c|}{}                         &                        & \multicolumn{1}{c|}{size}     & Binary Classification      & ROC-AUC & 26169 & 4112  & 3961  \\ \cline{2-8} 
\multicolumn{1}{c|}{}                         & \multirow{2}{*}{Motif} & \multicolumn{1}{c|}{basis}    & Multi-label Classification & ACC     & 18000 & 3000  & 3000  \\
\multicolumn{1}{c|}{}                         &                        & \multicolumn{1}{c|}{size}     & Multi-label Classification & ACC     & 18000 & 3000  & 3000  \\ \cline{2-8} 
\multicolumn{1}{c|}{}                         & CMNIST                 & \multicolumn{1}{c|}{color}    & Multi-label Classification & ACC     & 42000 & 7000  & 7000  \\ \cline{2-8} 
\multicolumn{1}{c|}{}                         & \multirow{2}{*}{PCBA} & \multicolumn{1}{c|}{scaffold}    & Multi-task Binary Classification & AP     & 262764 & 44019 & 43562  \\
\multicolumn{1}{c|}{}                         &                        & \multicolumn{1}{c|}{size}     & Multi-task Binary Classification & AP    & 269990 & 43792 & 31925  \\  \midrule
\multicolumn{1}{c|}{\multirow{6}{*}{DrugOOD}} & \multirow{3}{*}{IC50}  & \multicolumn{1}{c|}{assay}    & Binary Classification      & ROC-AUC & 34953 & 19475 & 19463 \\
\multicolumn{1}{c|}{}                         &                        & \multicolumn{1}{c|}{scaffold} & Binary Classification      & ROC-AUC & 22025 & 19478 & 19480 \\
\multicolumn{1}{c|}{}                         &                        & \multicolumn{1}{c|}{size}     & Binary Classification      & ROC-AUC & 37497 & 17987 & 16761 \\ \cline{2-8} 
\multicolumn{1}{c|}{}                         & \multirow{3}{*}{EC50}  & \multicolumn{1}{c|}{assay}    & Binary Classification      & ROC-AUC & 4978  & 2761  & 2725  \\
\multicolumn{1}{c|}{}                         &                        & \multicolumn{1}{c|}{scaffold} & Binary Classification      & ROC-AUC & 2743  & 2723  & 2762  \\
\multicolumn{1}{c|}{}                         &                        & \multicolumn{1}{c|}{size}     & Binary Classification      & ROC-AUC & 5189  & 2495  & 2505  \\ \bottomrule
\end{tabular}%
}\label{data_st}
\vspace{-4mm}
\end{table}
\subsection{Experimental Setup}\label{subsec:setup}
\noindent \textbf{Datasets.} We evaluate the performance of \ourmethod on two real-world benchmarks, GOOD~\cite{good} and DrugOOD~\cite{drugood}, with various distribution shifts to evaluate our method. Specifically, GOOD is a comprehensive graph OOD benchmark, and we selected three datasets: (1) GOOD-HIV~\cite{wu2018moleculenet}, a molecular graph dataset predicting HIV inhibition; (2) GOOD-CMNIST~\cite{arjovsky2019invariant}, containing graphs transformed from MNIST using superpixel techniques; (3) GOOD-Motif~\cite{wu2022discovering}, a synthetic dataset where graph motifs determine the label; and (4) GOOD-PCBA~\cite{wu2018moleculenet}, is a real-world molecular dataset which  includes 128 bioassays, forming 128 binary classification tasks. Due to the extremely unbalanced classes (only $1.4\%$ positive labels), we use the Average Precision (AP) averaged over the tasks as the evaluation metric. DrugOOD is designed for AI-driven drug discovery with three types of distribution shifts: scaffold, size, and assay, and applies these to two measurements (IC50 and EC50).
We detail various types of distribution-splitting strategies for different datasets.
\begin{itemize}
\item \textbf{Scaffold.}  Molecular scaffold is the core structure of a molecule that supports its overall composition, but it only exhibits specific properties when combined with particular functional groups. 
\item \textbf{Size.} The size of a graph refers to the total number of nodes, and it is also implicitly related to the graph's structural properties.     
\item \textbf{Assay.} The assay is an experimental technique used to examine or determine molecular characteristics. Due to differences in assay conditions and targets, activity values measured by different assays can vary significantly. 
\item \textbf{Basis.} The generation of a motif involves combining a base graph (wheel, tree, ladder, star, and path) with a motif (house, cycle, and crane), but only the motif is directly associated with the label. 
\item \textbf{Color.} CMNIST is a graph dataset constructed from handwritten digit images. Following previous research, we declare a distribution shift when the color of the handwritten digits changes.
\end{itemize}
We use the default dataset split proposed in each benchmark.  Statistics of each dataset are in Table \ref{data_st}.
\input{tables/main_good}
\input{tables/main_drugood_full}

\noindent\textbf{Baselines}.  We compare \ourmethod against ERM and two kinds of OOD baselines: (1)~Traditional OOD generalization approaches, including  Coral~\cite{coral}, IRM~\cite{arjovsky2019invariant} and VREx~\cite{krueger2021out}; (2)~graph-specific OOD generalization methods, including environment-based approaches (MoleOOD~\cite{yang2022learning}, CIGA~\cite{chen2022learning}, GIL~\cite{li2022learning}, and GREA~\cite{liu2022graph}, IGM~\cite{jia2024graph}), causal explanation-based approaches (Disc~\cite{fan2022debiasing} and DIR\cite{wu2022discovering}), and advanced architecture-based approaches (CAL~\cite{sui2022causal} and GSAT~\cite{miao2022interpretable}, iMoLD~\cite{zhuang2023learning}), GALA~\cite{Equad}, EQuAD~\cite{gala}. In our experiments, the methods we compared can be divided into two categories, one is ERM and traditional OOD generalization methods:
\begin{itemize}
    \item \textbf{ERM} is a standard learning approach that minimizes the average training error, assuming the training and test data come from the same distribution.
    \item \textbf{IRM}~\cite{arjovsky2019invariant} aims to learn representations that remain invariant across different environments, by minimizing the maximum error over all environments.
    \item \textbf{VREx}~\cite{krueger2021out} propose a penalty on the variance of training risks which can providing more robustness to changes in the input distribution.  
    \item \textbf{Coral}~\cite{coral} utilize a nonlinear transformation to align the second-order statistical features of the source and target domain distributions
\end{itemize}
Another class of methods is specifically designed for Graph OOD generalization:
\begin{itemize}
    \item \textbf{MoleOOD}~\cite{yang2022learning} learn the environment invariant molecular substructure by a environment inference model and a molecular decomposing model.
    \item \textbf{CIGA}~\cite{chen2022learning} proposes an optimization objective based on mutual information to ensure the learning of invariant subgraphs that are not affected by the environment.
    \item \textbf{GIL}~\cite{li2022learning} performs environment identification and invariant risk loss optimization by separating the invariant subgraph and the environment subgraph.
    \item \textbf{GERA}~\cite{liu2022graph} performs data augmentation by replacing the input graph with the environment subgraph to improve the generalization ability of the model
    \item \textbf{IGM}~\cite{jia2024graph} performs data augmentation by simultaneously performing a hybrid strategy of invariant subgraphs and environment subgraphs.
    \item \textbf{DIR}~\cite{wu2022discovering} identifies causal relation between input graphs and labels by performing counterfactual interventions.
    \item \textbf{DisC}~\cite{fan2022debiasing} learns causal and bias representations through a causal and disentangling based learning strategy separately.
    \item \textbf{GSAT}~\cite{miao2022interpretable} learns the interpretable label-relevant subgraph through an stochasticity attention mechanism.
    \item \textbf{CAL}~\cite{sui2022causal} proposes a causal attention learning strategy to ensure that GNNs learn effective representations instead of optimizing loss through shortcuts.
    \item  \textbf{iMoLD}~\cite{zhuang2023learning} designs two GNNs to directly extract causal features from the encoded graph representation.
    \item { \textbf{GALA}~\cite{gala} designs designs a new loss function to ensure graph OOD generalization without environmental information as much as possible.}
    \item  {\textbf{EQuAD}~\cite{Equad} learns how to effectively remove spurious features by optimizing the self-supervised informax function.}
    \item { \textbf{HSE}~\cite{piao2024improving} propose a hierarchical semantic environment generation method for graphs by extracting variant subgraphs and applying stochastic attention.}
    \item  {\textbf{LIRS}~\cite{yaolearning} indirectly learns graph-invariant features by first identifying and removing spurious features from those learned via ERM.}
\end{itemize}

\noindent\textbf{Implementation Details}. To ensure fairness, we adopt the same experimental setup as iMold across two benchmarks. For molecular datasets with edge features, we use a three-layer GIN with a hidden dimension of 300, while for non-molecular graphs, we employ a four-layer GIN with a hidden dimension of 128. The projector is a two-layer MLP with a hidden dimension set to half that of the GIN encoder. EMA rate $\alpha$ for prototype updating is fixed at 0.99. Adam optimizer is used for model parameter updates.  All baselines use the optimal parameters from their original papers. 

\subsection{Performance Comparison}\label{subsec:results}

In this experiment, we aim to answer
\textbf{Q1: Whether \ourmethod achieves the best performance on OOD generalization benchmarks?} 
The answer is \textbf{YES}, since \ourmethod shows the best results on the majority of datasets. Specifically, we have the following observations.

 \noindent$\rhd$ \textsf{State-of-the-art results.}
According to Table~\ref{tab:merged_good} and ~\ref{tab:main_drugood}, \ourmethod achieves state-of-the-art performance on 11 out of 13 datasets, and secures the second place on the remaining dataset. The average improvements against the previous SOTA are $2.17\%$ on GOOD and $1.68\%$ on DrugOOD. Notably, \ourmethod achieves competitive performance across various types of datasets with different data shifts, demonstrating its generalization ability on different data. Moreover, our model achieves the best results in both binary and multi-class tasks, highlighting the effectiveness of the multi-prototype classifier in handling different classification tasks.

 \noindent$\rhd$ \textsf{Sub-optimal performance of environment-based methods.}
Among all baselines, environment-based methods only achieve the best performance on 3 datasets, while architecture-based OOD generalization methods achieve the best results on most datasets. These observations suggest that environment-based methods are limited by the challenge of accurately capturing environmental information in graph data, leading to a discrepancy between theoretical expectations and empirical results. In contrast, the remarkable performance of \ourmethod also proves that graph OOD generalization can still be achieved without specific environmental information.

\subsection{Ablation Study}
We aim to discover 
\textbf{Q2: Does each module in \ourmethod contribute to effective OOD generalization?} The answer is \textbf{YES}, as removing any key component leads to performance degradation, as demonstrated by the results in Table~\ref{tab:ablation}. We have the following discussions.

\input{tables/ablation}
\noindent$\rhd$ \textsf{Ablation on $\mathcal{L}_{\mathrm{IPM}}$ and $\mathcal{L}_{\mathrm{PS}}$.}
We remove $\mathcal{L}_{\mathrm{IPM}}$ and $\mathcal{L}_{\mathrm{PS}}$ in the Eq.~\eqref{eq: target2} respectively to explore their impacts on the performance of graph OOD generalization. The experimental results demonstrate a clear fact: merely optimizing for invariance (w/o $\mathcal{L}_{\mathrm{PS}}$) or separability (w/o $\mathcal{L}_{\mathrm{IPM}}$) weakens the OOD generalization ability of our model, especially for the multi-class classification task, as shown in CMNIST-color. This provides strong evidence that ensuring both invariance and separability is a sufficient and necessary condition for effective OOD generalization in graph learning.
\noindent$\rhd$ \textsf{Ablation on the design of \ourmethod.} To verify the effectiveness of each module designed for \ourmethod,
we conducted ablation studies by removing the hyperspherical projection(w/o Project), multi-prototype mechanism (w/o Multi-P), invariant encoder (w/o Inv.Enc), and prototype-related weight calculations (w/o Update) and pruning (w/o Prune). The results confirm their necessity. First, removing the hyperspherical projection significantly drops performance, as optimizing Eq.~(\ref{eq: target2}) requires hyperspherical space. Without it, results are even worse than ERM. Similarly, setting the prototype count to one blurs decision boundaries and affects the loss function $\mathcal{L}_{\mathrm{PS}}$, compromising inter-class separability. Lastly, replacing the invariant encoder $\mathrm{GNN}_{S}$ with $\mathrm{GNN}_{E}$ directly introduces environment-related noise, making it difficult to obtain effective invariant features, thus hindering OOD generalization.
Additionally, the removal of prototype-related weight calculations and weight pruning degraded prototypes into the average of all class samples, resulting in the prototypes degrading into the average representation of all samples in the class, failing to maintain classification performance in OOD scenarios.
\begin{figure*}[!t]
    \centering
    \subfigure[Sensitivity of $k$]{ \label{subfig:paramk}
    \includegraphics[height=0.16\textwidth]{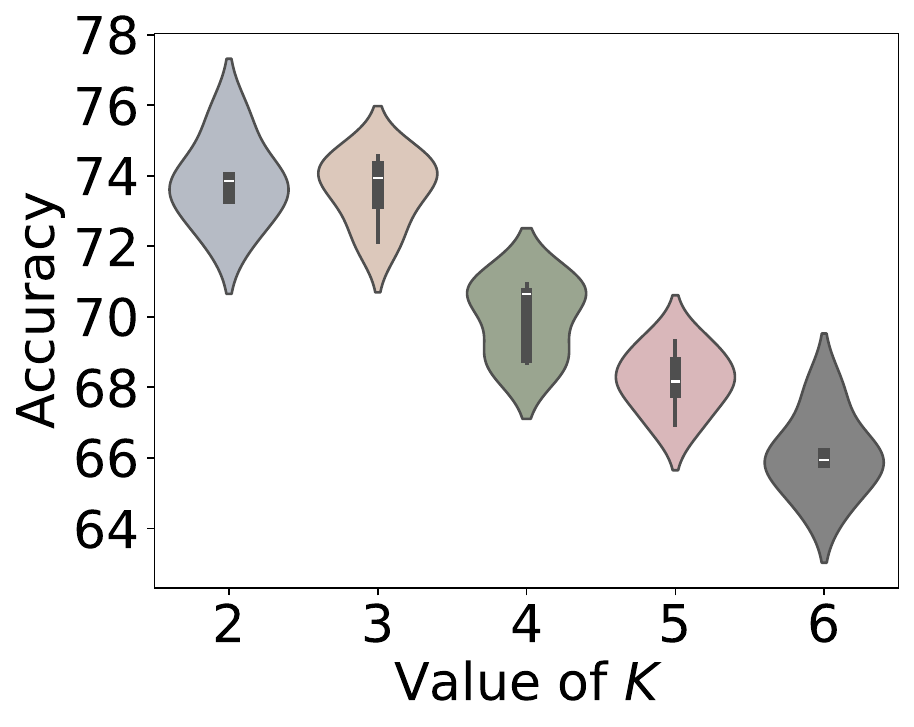}
    }
    \hfill
    \subfigure[Sensitivity of $\beta$]{ \label{subfig:paramb}
    \includegraphics[height=0.16\textwidth]{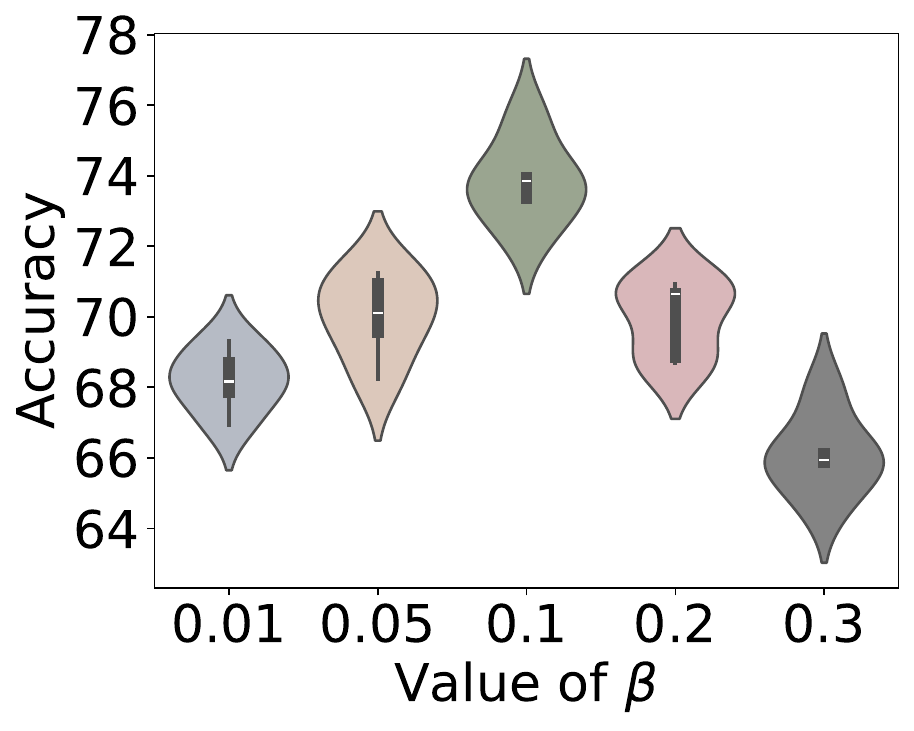}
    }
    \hfill
    \subfigure[Impact of prototype updating mechanisms]{ \label{subfig:update}
    \includegraphics[height=0.16\textwidth]{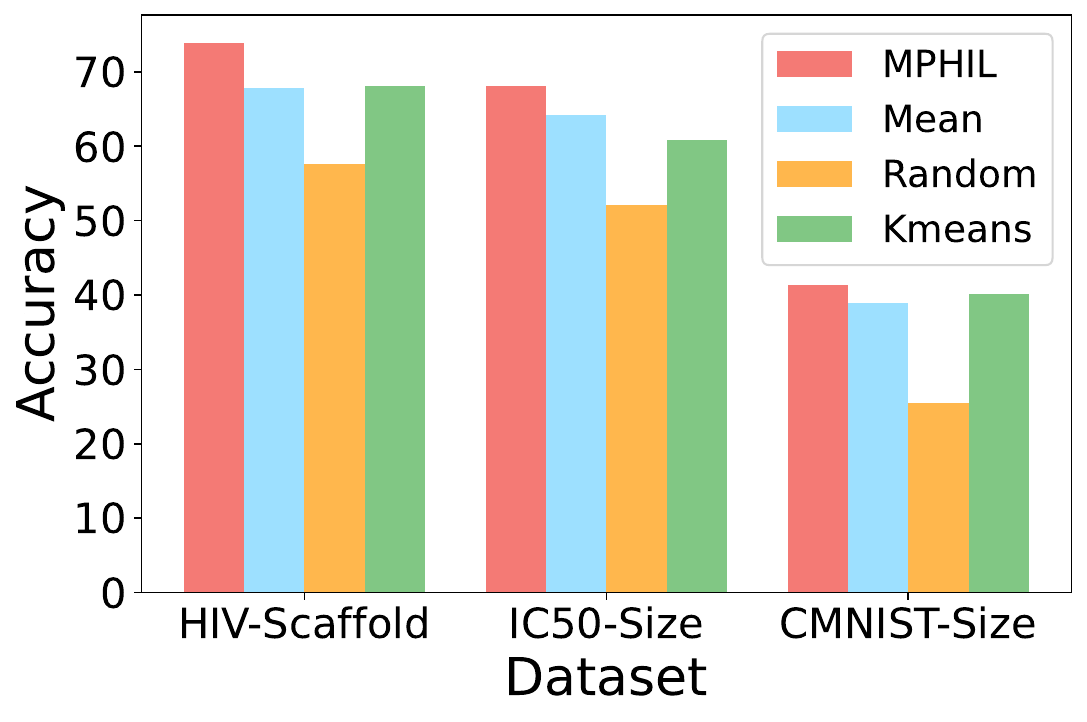}
    }
    \hfill
    \subfigure[Impact of different statistical metrics]{ \label{subfig:metric}
    \includegraphics[height=0.16\textwidth]{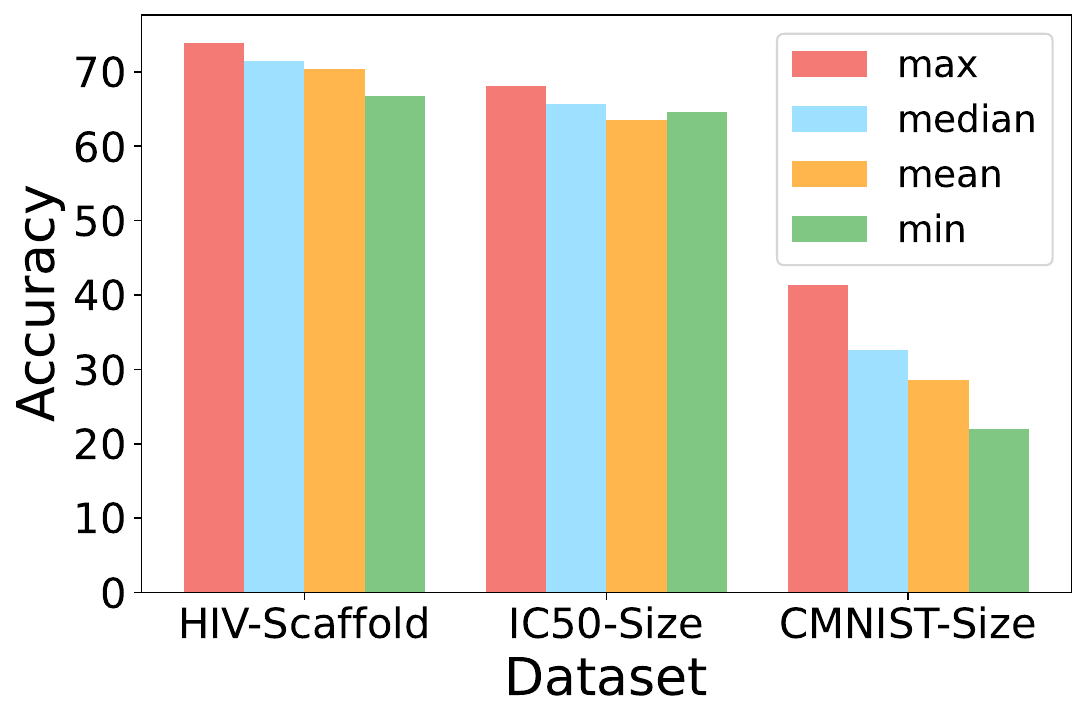}
    }
    
    \caption{The two figures on the left present a hyperparameter analysis of the  $K$ and $\beta$, while the two figures on the right illustrate the comparison of different module designs on prototype update and metric used in Eq.~\eqref{classify}. } 
    
    \label{fig:four_images}
    
\end{figure*}
\subsection{In-Depth Analysis}
In this experiment, we will investigate \textbf{Q3: How do the details (hyperparameter settings and variable designs) in \ourmethod impact performance?} The following experiments are conducted to answer this question and the experimental results are in Fig.~\ref{fig:four_images}.

\noindent$\rhd$ \textsf{Hyperparameter Analysis.}
To investigate the sensitivity of the number of prototypes and the coefficient $\beta$ in $\mathcal{L}_{\mathrm{IPM}}$ on performance, we vary $k$ from 2 to 6 and $\beta$ from $\{0.01, 0.05, 0.1, 0.2, 0.3\}$. Our conclusions are as follows: \ding{192} In Fig.~\ref{subfig:paramk}, the best performance is achieved when the number of prototypes is approximately twice the number of classes. Deviating from this optimal range, either too many or too few prototypes negatively impacts the final performance. \ding{193} According to Fig.~\ref{subfig:paramb}, a smaller $\beta$  hampers the model’s ability to effectively learn invariant features, while selecting a moderate $\beta$ leads to the best performance.
 

\noindent$\rhd$ \textsf{Module design analysis.}
To investigate the impact of different prototype update mechanisms and statistical metrics in Eq.~\eqref{classify}, we conducted experimental analyses and found that \ding{192} According to Fig.~\ref{subfig:update}, all compared methods lead to performance drops due to their inability to ensure that the updated prototypes possess both intra-class diversity and inter-class separability, which is the key to the success of MPHIL's prototype update method. \ding{193} In Fig.~\ref{subfig:metric}, $\max$ achieves the best performance by selecting the most similar prototype to the sample, helping the classifier converge faster to the correct decision space.
\begin{figure}[t]

\centering 
\subfigure[1-order Wasserstein distance]{ \label{wl_Dis}
\includegraphics[width=0.17\textwidth,height=0.13\textwidth]{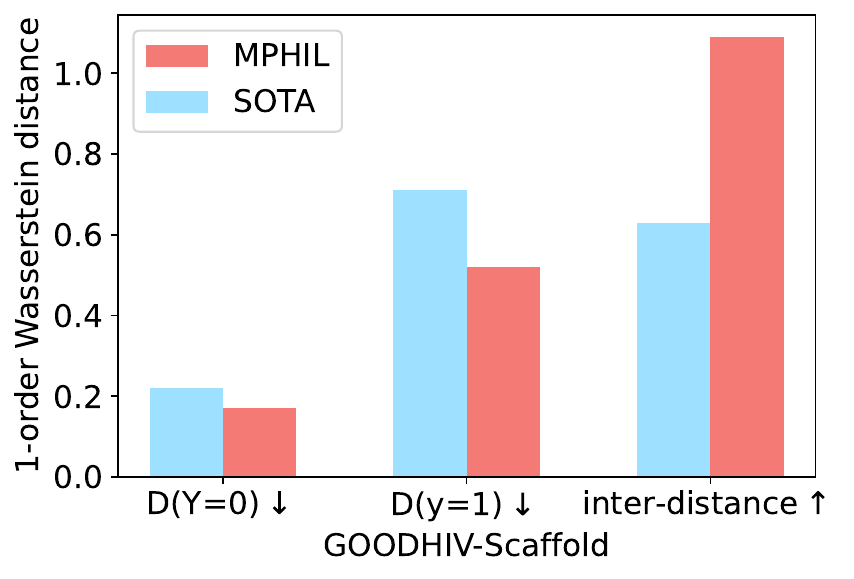}
}
\subfigure[T-SNE visualization, left: \ourmethod, right: SOTA ]{ \label{tsne}
\includegraphics[width=0.13\textwidth,height=0.13\textwidth]{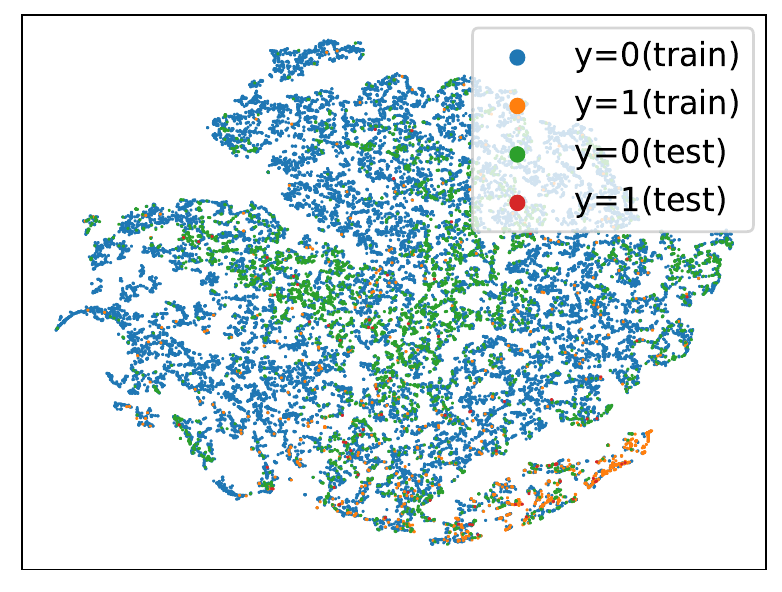}
\includegraphics[width=0.13\textwidth,height=0.13\textwidth]{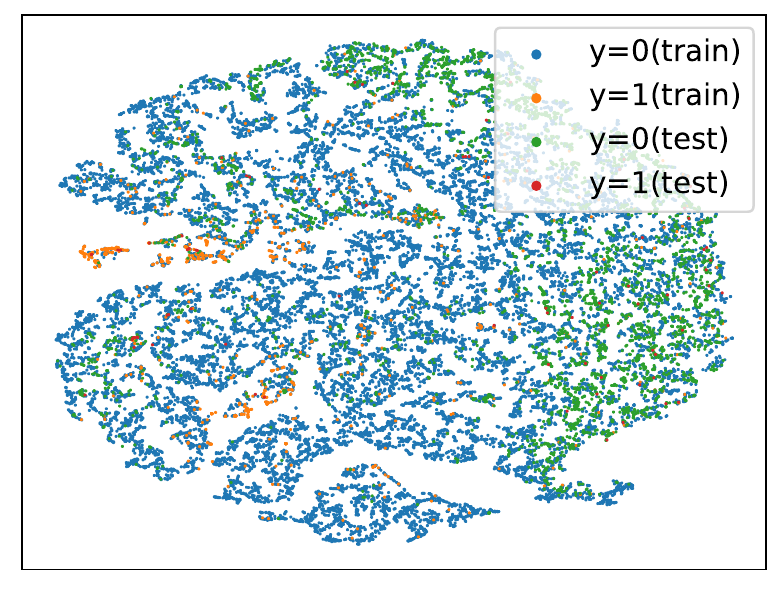}
}

\caption{Visualization and quantitative analysis of the separability advantages in hyperspherical space on HIV-scaffold.}

\end{figure}



\begin{figure}[t]
    \centering

    \includegraphics[width=0.45\textwidth]{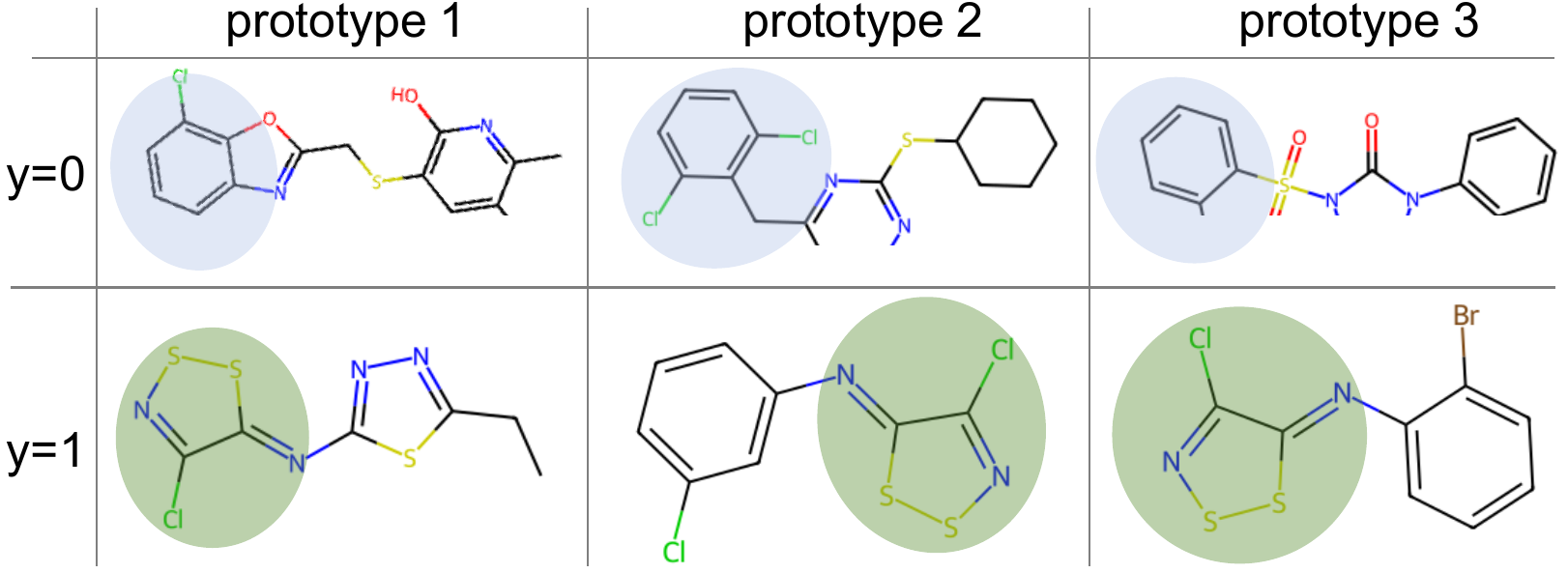}

    \caption{Visualizations of prototypes and invariant subgraphs~(highlighted) of IC50-assay dataset.}
    \label{fig:proto}

\end{figure}

\begin{table}[htbp]
\centering
\caption{Performance comparison on SPMotif-binary dataset with varying bias levels}
\label{tab:spmotif_results}
\begin{tabular}{lccc}
\toprule
\textbf{Method} & \multicolumn{3}{c}{\textbf{SPMotif-binary}} \\
\cmidrule(lr){2-4}
 & $b = 0.40$ & $b = 0.60$ & $b = 0.90$ \\
\midrule
ERM    & 74.93 $\pm$ 3.94 & 72.78 $\pm$ 3.15 & 63.78 $\pm$ 4.18 \\
IRM    & 76.01 $\pm$ 4.12 & 70.85 $\pm$ 4.73 & 66.55 $\pm$ 4.80 \\
VREx   & 79.03 $\pm$ 1.02 & 73.78 $\pm$ 1.75 & 65.27 $\pm$ 6.78 \\ \midrule
GIL    & 77.15 $\pm$ 3.18 & 73.85 $\pm$ 2.76 & 68.90 $\pm$ 7.28 \\
GREA   & 79.65 $\pm$ 6.36 & 73.01 $\pm$ 7.99 & 69.85 $\pm$ 0.35 \\
CIGA   & 76.93 $\pm$ 3.94 & 71.70 $\pm$ 1.55 & 66.80 $\pm$ 5.35 \\
AIA    & 78.46 $\pm$ 3.19 & 71.83 $\pm$ 0.69 & 64.37 $\pm$ 4.14 \\
EQuAD  & 80.82 $\pm$ 0.65 & 74.20 $\pm$ 4.10 & 69.79 $\pm$ 7.81 \\ 
LIRS   & {82.17} $\pm$ 0.91 & {75.32} $\pm$ 1.65 & {71.29} $\pm$ 2.12 \\ \midrule
\ourmethod   & \textbf{83.26} $\pm$ 0.74 & \textbf{76.04} $\pm$ 1.58 & \textbf{72.14} $\pm$ 1.48\\ 
\bottomrule
\end{tabular}
\end{table}
\subsection{Visualized Validation}
In this subsection, we aim to investigate \textbf{Q4: Can these key designs (i.e., hyperspherical space and multi-prototype mechanism) tackle two unique challenges in graph OOD generalization tasks?} The answer is \textbf{YES}, we conduct the following visualization experiments to verify this conclusion. 

\noindent$\rhd$ \textsf{Hyperspherical representation space.} To validate the advantage of hyperspherical space in enhancing class separability, we compare the 1-order Wasserstein distance~\cite{villani2009optimal} between same-class and different-class samples, as shown in Fig.~\ref{wl_Dis}.
It is evident that \ourmethod produces more separable invariant representations (higher inter-class distance), while also exhibiting tighter clustering for samples of the same class (lower intra-class distance). In contrast, although traditional latent spaces-based SOTA  achieves a certain level of intra-class compactness, its lower separability hinders its overall performance. Additionally, we visualized the sample representations learned by our \ourmethod and SOTA using t-SNE in Fig.~\ref{tsne}, where corresponding phenomenon can be witnessed.

\noindent$\rhd$ \textsf{Prototypes visualization.} We also reveal the characteristics of prototypes by visualizing samples that exhibit the highest similarity to each prototype. Fig.~\ref{fig:proto} shows that prototypes from different classes capture distinct invariant subgraphs, ensuring a strong correlation with their respective labels. Furthermore, within the same category, different prototypes encapsulate samples with varying environmental subgraphs. This validates that multi-prototype learning can effectively capture label-correlated invariant features without explicit environment definitions, which solve the challenges of out-of-distribution generalization in real-world graph data.
\subsection{Analysis of Spurious Change}
In this section, we will investigate \textbf{Q5: Whether \ourmethod can maintain consistently strong performance under varying degrees of distribution shifts.} The answer is \textbf{YES}. To validate this, we conduct experiments on the SPMotif-binary dataset, which is constructed following the setting in LIRS~\cite{yaolearning}. In this dataset, the correlation strength between spurious subgraphs and class labels is controlled by a bias parameter $b$, allowing us to systematically evaluate model robustness under different levels of spurious correlations. As shown in Table~\ref{tab:spmotif_results}, our method consistently outperforms all baseline approaches including general OOD methods like IRM and VREx, as well as graph-specific methods such as GIL, CIGA, and EQuAD, across all bias levels. Notably, while LIRS achieves strong performance by indirectly removing spurious features, our method further improves upon it, especially under high spurious correlation ($b = 0.90$), suggesting enhanced capability in identifying and preserving invariant features. These results demonstrate the superior OOD generalization ability of \ourmethod, showing consistently strong performance under varying levels of spurious correlations and distribution shifts.
\begin{figure}[t]

\centering 
\subfigure[Execution time]{ \label{train_time}
\includegraphics[width=0.22\textwidth,height=0.15\textwidth]{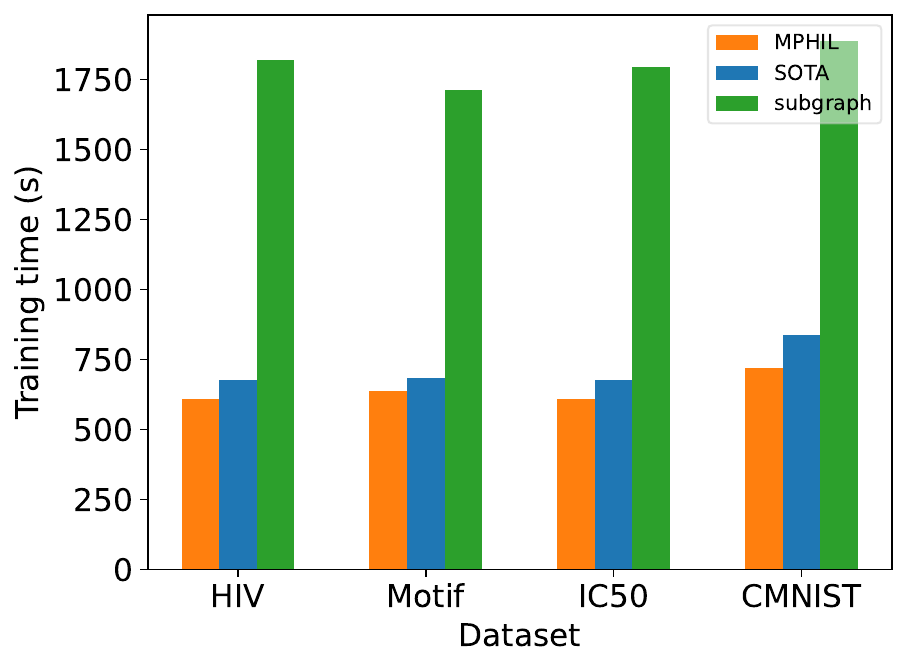}
}
\subfigure[Execution time under varying class]{ \label{class}
\includegraphics[width=0.22\textwidth,height=0.15\textwidth]{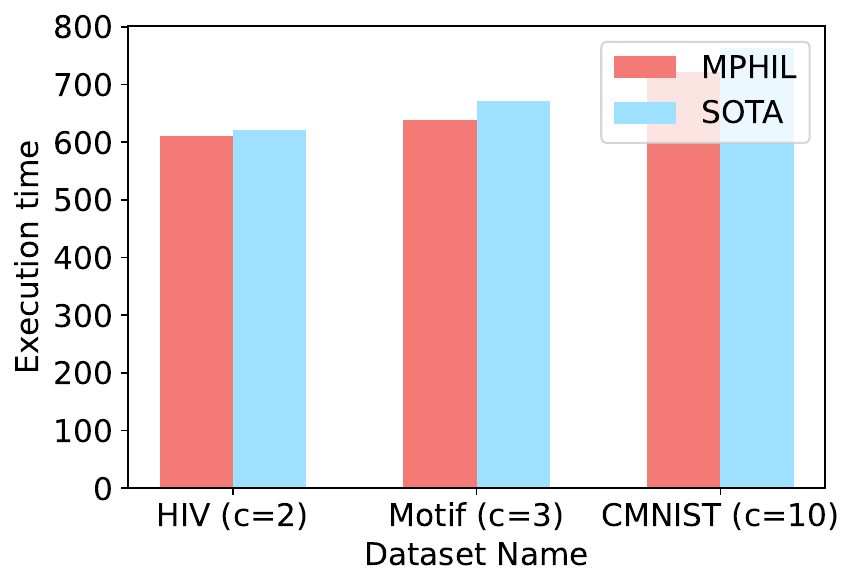}
}

\caption{Comparison of execution times of different methods under different datasets.}

\end{figure}
\subsection{Computational Complexity Comparison}
In this section, we will investigate \textbf{Q6: Does \ourmethod also offer advantages in computational efficiency?} The answer is \textbf{YES}. we provide empirical evidence through runtime analysis, as shown in Fig~\ref{train_time} and ~\ref{class}.

\noindent$\rhd$ \textsf{Efficiency on execution time.}
To assess the runtime efficiency of our method, we compare it with both subgraph-based and other SOTA methods across four representative datasets, as shown in Fig.~\ref{train_time}. Our method consistently achieves the best execution time. In particular, compared with subgraph-based approaches, which require costly operations to extract invariant subgraphs, our method avoids this overhead by conducting invariant learning directly in the hyperspherical space. This design leads to a substantial reduction in computation time. Furthermore, our method outperforms other non-subgraph-based SOTA method in terms of efficiency, demonstrating both its scalability and practicality in real-world applications.

\noindent$\rhd$ \textsf{Efficiency with Number of Classes.} Our proposed \ourmethod performs hyperspherical invariant representation extraction in the latent space and classifies samples by measuring distances to class prototypes on the Hyperspherical space.
As the number of classes increases, our method introduces additional prototypes accordingly. However, this introduces only minimal computational overhead. As shown in the Fig~\ref{class}, moving from 2 to 10 classes results in only a slight increase in execution time (e.g., 610s for HIV with 2 classes to 721s for CMNIST with 10 classes), demonstrating the scalability of our approach with respect to class size.


%% file: tables/main_good.tex
\begin{table*}[h!]
\caption{Performance comparison in terms of average accuracy (standard deviation) on GOOD benchmark. The best and runner-up results are highlighted in \textbf{bold} and \underline{underlined}.}
\centering
\resizebox{\textwidth}{!}{
\begin{tabular}{l|cc|c|cc|cc}
\toprule
\multirow{2}{*}{\textbf{Method}} & \multicolumn{2}{c|}{\textbf{GOOD-Motif}} & \multicolumn{1}{c|}{\textbf{GOOD-CMNIST}} & \multicolumn{2}{c|}{\textbf{GOOD-HIV}} & \multicolumn{2}{c}{\textbf{GOOD-PCBA}}  \\
 & \textbf{basis} & \textbf{size} & \textbf{color} & \textbf{scaffold} & \textbf{size} & \textbf{scaffold} & \textbf{size}  \\
\midrule
ERM & 60.93 (2.11) & 46.63 (7.12) & 26.64 (2.37) & 69.55 (2.39) & 59.19 (2.29) & 21.93 (0.74) & 15.60 (0.55)  \\
IRM & 64.94 (4.85) & {54.52 (3.27)} & 29.63 (2.06) & 70.17 (2.78) & 59.94 (1.59) & 22.37 (0.43) & 15.82 (0.52) \\
VREX & 61.59 (6.58) & {55.85 (9.42)} & 27.13 (2.90) & 69.34 (3.54) & 58.49 (2.28) & 21.65 (0.82) & 15.85 (0.47) \\
Coral & 61.95 (4.36) & 55.80 (4.05) & 29.21 (6.87) & 70.69 (2.25) & 59.39 (2.90) & 22.00 (0.46) & 16.88 (0.58)  \\
\midrule
MoleOOD & - & - & - & 69.39 (3.43) & 58.63 (1.78) & 12.92 (1.23) & 10.30 (0.36)  \\
CIGA & 67.81 (2.42) & 51.87 (5.15) & 25.06 (3.07) & 69.40 (1.97) & {61.81 (1.68)} & - & -  \\
GIL & 65.30 (3.02) & 54.65 (2.09) & 31.82 (4.24) & 68.59 (2.11) & 60.97 (2.88) & - & - \\
GREA & 59.91 (2.74) & 47.36 (3.82) & 22.12 (5.07) & {71.98 (2.87)} & 60.11 (1.07) & 20.23 (1.53) & 13.82 (1.18) \\
IGM & {74.69 (8.51)} & 52.01 (5.87) & 33.95 (4.16) & {71.36 (2.87)} & 62.54 (2.88) & 20.09 (1.99) & 12.87 (0.69)\\
\midrule
DIR & 64.39 (2.02) & 43.11 (2.78) & 22.53 (2.56) & 68.44 (2.51) & 57.67 (3.75) & 23.82 (0.89) & 16.80 (1.17) \\
DisC & 65.08 (5.06) & 42.23 (4.20) & 23.53 (0.67) & 58.85 (7.26) & 49.33 (3.84) & - & -  \\
\midrule
GSAT & 62.27 (8.79) & 50.03 (5.71) & {35.02 (2.78)} & 70.07 (1.76) & 60.73 (2.39) & 20.18 (0.74) & 13.52 (0.90) \\
CAL & {68.01 (3.27)} & 47.23 (3.01) & 27.15 (5.66) & 69.12 (1.10) & 59.34 (2.14) & 18.62 (1.21) & 13.01 (0.65)  \\
iMoLD & - & - & - & \underline{72.05 (2.16)} & {62.86 (2.34)} & 22.58 (0.67) & 18.21 (1.10) \\
\textcolor{black}{GALA} & \textcolor{black}{72.97 (4.28)} & \textcolor{black}{\textbf{60.82 (0.51)}} & \textcolor{black}{\underline{40.62 (2.11)}} & \textcolor{black}{71.22 (1.93)} & \textcolor{black}{\underline{65.29 (0.72)}} & 21.26 (1.48) & 13.57 (0.88)  \\
\textcolor{black}{EQuAD} & \textcolor{black}{\underline{75.46 (4.35)}} & \textcolor{black}{55.10 (2.91)} & \textcolor{black}{{40.29 (3.95)}} & \textcolor{black}{71.49 (0.67)} & \textcolor{black}{{64.09 (1.08)}} & \textcolor{black}{\underline{23.17 (0.80)}} & \textcolor{black}{17.61 (0.76)} \\
\textcolor{black}{HSE} & \textcolor{black}{74.62 (3.05)} & \textcolor{black}{54.03 (1.85)} & \textcolor{black}{39.17 (5.17)} & \textcolor{black}{70.85 (1.97)} & \textcolor{black}{63.41 (2.14)} & 17.41 (0.83) & 17.53 (0.76)  \\
\textcolor{black}{LIRS} & \textcolor{black}{74.31 (4.66)} & \textcolor{black}{53.78 (2.17)} & \textcolor{black}{39.02 (3.78)} & \textcolor{black}{70.54 (0.51)} & \textcolor{black}{63.10 (1.21)} & 20.98 (1.57) & 19.11 (1.34)  \\
\midrule
\ourmethod & \textbf{76.23 (4.89)} & \underline{58.43 (3.15)} & \textbf{41.29 (3.85)} & \textbf{73.94 (1.77)} & \textbf{66.84 (1.09)} & \textbf{25.73 (0.79)} & \textbf{{19.36 (1.06)}} \\
\bottomrule
\end{tabular}
\label{tab:merged_good}
}
\end{table*}

%% file: tables/main_drugood_full.tex
\begin{table*}[h!] 
\caption{Performance comparison in terms of average accuracy (standard deviation) on DrugOOD benchmark. The best and runner-up results are highlighted in \textbf{bold} and \underline{underlined}.}
\centering
\resizebox{0.95\textwidth}{!}{
\begin{tabular}{l|ccc|ccc}
\toprule
\multirow{2}{*}{\textbf{Method}} & \multicolumn{3}{c|}{\textbf{DrugOOD-IC50}} & \multicolumn{3}{c}{\textbf{DrugOOD-EC50}} \\ 
 & \textbf{assay} & \textbf{scaffold} & \textbf{size} & \textbf{assay} & \textbf{scaffold} & \textbf{size} \\ \midrule
ERM & 70.61 (0.75) & 67.54 (0.42) & 66.10 (0.31) & 65.27 (2.39) & 65.02 (1.10) & 65.17 (0.32) \\ 
IRM & 71.15 (0.57) & 67.22 (0.62) & \underline{67.58 (0.58)} & 67.77 (2.71) & 63.86 (1.36) & 59.19 (0.83) \\ 
VREx & 70.98 (0.77) & 68.02 (0.43) & 65.67 (0.19) & 69.84 (1.88) & 62.31 (0.96) & {65.89 (0.83)} \\ 
Coral & 71.28 (0.91) & 68.36 (0.61) & 67.53 (0.32) & 72.08 (2.80) & 64.83 (1.64) & 58.47 (0.43) \\
G-mixup  & 70.33 (0.78) & 67.91 (0.46) & 66.58 (0.29) & 64.88 (2.30) & 65.77 (1.12) & 65.49 (0.34) \\\midrule
MoleOOD & 71.62 (0.50) & 68.58 (1.14) & 67.22 (0.96) & 72.69 (4.16) & 65.78 (1.47) & 64.11 (1.04) \\ 
CIGA & \underline{71.86 (1.37)} & \textbf{69.14 (0.70)} & 66.99 (1.40) & 69.15 (5.79) & \underline{67.32 (1.35)} & 65.60 (0.82) \\ 
GIL & 70.66 (1.75) & 67.81 (1.03) & 66.23 (1.98) & 70.25 (5.79) & 63.95 (1.17) & 64.91 (0.76) \\ 
GREA & 70.23 (1.17) & 67.20 (0.77) & 66.09 (0.56) & 74.17 (1.47) & 65.84 (1.35) & 61.11 (0.46) \\ 
IGM & 68.05 (1.84) & 63.16 (3.29) & 63.89 (2.97) & 76.28 (4.43) & 67.57 (0.62) & 60.98 (1.05) \\ \midrule
DIR & 69.84 (1.41) & 66.33 (0.65) & 62.92 (1.89) & 65.81 (2.93) & 63.76 (3.22) & 61.56 (4.23) \\ 
DisC & 61.40 (2.56) & 62.70 (2.11) & 64.43 (0.60) & 63.71 (5.56) & 60.57 (2.27) & 57.38 (2.48) \\ \midrule
GSAT & 70.59 (0.43) & 66.94 (1.43) & 64.53 (0.51) & 73.82 (2.62) & 62.65 (1.79) & 62.65 (1.79) \\ 
CAL & 70.09 (1.03) & 65.90 (1.04) & 64.42 (0.50) & {74.54 (1.48)} & 65.19 (0.87) & 61.21 (1.76) 
\\
iMoLD & 71.77 (0.54) & 67.94 (0.59) & 66.29 (0.74) & {77.23 (1.72)} & 66.95 (1.26) & {67.18 (0.86)} 
\\ 

\textcolor{black}{GALA} & \textcolor{black}{70.58 (2.63)} & \textcolor{black}{66.35 (0.86)} & \textcolor{black}{66.54 (0.93)}  & \textcolor{black}{{77.24 (2.17)}} & \textcolor{black}{66.98 (0.84)} & \textcolor{black}{63.71 (1.17)}  \\
\textcolor{black}{EQuAD} &\textcolor{black}{71.57 (0.95)} & \textcolor{black}{67.74 (0.57)} & \textcolor{black}{67.54 (0.27)} &\textcolor{black}{\underline{77.64 (0.63)}} &\textcolor{black}{65.73 (0.17)} &\textcolor{black}{64.39 (0.67)}  \\
HSE & 69.53 (1.21) & 64.32 (1.18) & 66.37 (0.53) & {73.89 (2.07)} & 66.90 (1.72) & 62.94 (2.02) 
\\
LIRS & 71.06 (1.14) & 66.83 (1.02) & 65.72 (0.48) & {77.15 (1.58)} & 64.36 (1.61) & \underline{67.54 (0.66)} 
\\ 
\midrule
\ourmethod & \textbf{72.96 (1.21)} & \underline{68.62 (0.78)} & \textbf{68.06 (0.55)} & \textbf{78.08 (0.54)} & \textbf{68.34 (0.61)} & \textbf{68.11 (0.58)} \\

\bottomrule
\end{tabular}
}\label{tab:main_drugood}
\end{table*}

%% file: tables/ablation.tex
\begin{table}
\caption{Performance of \ourmethod and its variants.} 

\resizebox{1\columnwidth}{!}{
  \begin{tabular}{l|ccc}
    \toprule
    \textbf{Variants} &\textbf{CMNIST-color} &\textbf{HIV-scaffold} &\textbf{IC50-size} \\
    \midrule
\ourmethod & \textbf{41.29 (3.85)} & \textbf{74.69 (1.77)} & \textbf{68.06 (0.55)}  \\
    ERM & 26.64 (2.37) & 69.55 (2.39) & 66.10 (0.31)  \\
\midrule
w/o $\mathcal{L}_\textrm{IPM}$ & 37.86 (3.44) & 70.61 (1.52) & 67.09 (0.65)  \\
w/o $\mathcal{L}_\textrm{PS}$ & 37.53 (2.18) & 71.06 (1.56) & 66.21 (0.37) \\
\midrule
w/o Project & 21.05 (4.89) & 65.78 (3.57) & 51.96 (2.54) \\
w/o Multi-P & 20.58 (3.78) & 62.11 (1.95) & 57.64 (1.02)  \\
w/o {Inv. Enc.} & 34.86 (2.92) & 66.72 (1.19) & 63.73 (0.89)  \\
\midrule
w/o Update & 38.95 (3.01) & 67.89 (1.84) & 64.14 (1.22)  \\
w/o Prune & 40.58 (3.78) & 71.11 (1.95) & 67.64 (1.02) \\
    \bottomrule
  \end{tabular}}\label{tab:ablation}

\end{table}

%% file: 5_conclusion/conclusion.tex
In this work, we introduce a novel graph invariant learning framework integrated with hyperspherical space and prototypical learning, ensuring that the learned representations are both environment-invariant and class-separable without relying on environmental information. Building upon this framework, we present a new graph out-of-distribution generalization method named \ourmethod. \ourmethod achieves inter-class invariance and intra-class separability by optimizing two effective loss functions and leverages class prototypes, defined as the mean feature vectors of each category, to eliminate dependency on individual prototypes. Experimental evaluations on benchmarks demonstrate the effectiveness of \ourmethod.